\documentclass{article}

\usepackage{arxiv}

\usepackage{graphicx}

\usepackage[T1]{fontenc}
\usepackage[utf8]{inputenc}

\usepackage{amsmath,amsfonts}
\usepackage{nicefrac}

\usepackage{booktabs}

\usepackage[numbers]{natbib}
\usepackage{doi}

\usepackage{hyperref}
\usepackage{url}

\usepackage{cleveref}

\usepackage{lipsum}

\title{Robust Assembly Progress Estimation\\ via Deep Metric Learning}

\newif\ifuniqueAffiliation
\uniqueAffiliationtrue

\ifuniqueAffiliation 
\author{Kazuma Miura \\
The Precision Engineering Course, Graduate School of Science and Engineering\\
Chuo University\\
Tokyo, Japan\\
\texttt{miura@sensor.mech.chuo-u.ac.jp}
\And
Sarthak Pathak \\
The Department of Precision Mechanics, Faculty of Science and Engineering\\
Chuo University\\
Tokyo, Japan
\And
Kazunori Umeda \\
The Department of Precision Mechanics, Faculty of Science and Engineering\\
Chuo University\\
Tokyo, Japan \\
}
\else
\usepackage{authblk}

\newbox{\orcid}\sbox{\orcid}{\includegraphics[scale=0.06]{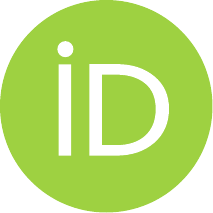}} 
\author[1]{%
	{\usebox{\orcid}\hspace{1mm}Kazuma Miura\thanks{\texttt{miura@sensor.mech.chuo-u.ac.jp}}%
}
\author[2]{%
	\href{https://orcid.org/0000-0002-5271-1782}{\usebox{\orcid}\hspace{1mm}Sarthak Pathak\thanks{}}%
}

\author[3]{%
	\href{https://orcid.org/0000-0002-4458-4648}{\usebox{\orcid}\hspace{1mm}Kazunori Umeda\thanks{}}%
}

\affil[1]{The Precision Engineering Course, Graduate School of Scienceand Engineering, Chuo University,Tokyo,Japan}
\affil[2]{The DepartmentofPrecision Mechanics,Faculty of Scienceand Engineering, Chuo University,Tokyo,Japan}
\affil[3]{The DepartmentofPrecision Mechanics,Faculty of Scienceand Engineering, Chuo University,Tokyo,Japan}
\fi

\renewcommand{\shorttitle}{\textit{arXiv} Template}

\hypersetup{
pdftitle={A template for the arxiv style},
pdfsubject={q-bio.NC, q-bio.QM},
pdfauthor={Kazuma Miura, Sarthak Pathak, Kazunori Umeda},
pdfkeywords={Deep metric learning, Quadruplet loss, Progress recognition, Assembly},
}

\begin{document}
\maketitle

\begin{abstract}
	In recent years, the advancement of AI technologies has accelerated the development of smart factories. In particular, the automatic monitoring of product assembly progress is crucial for improving operational efficiency, minimizing the cost of discarded parts, and maximizing factory productivity. However, in cases where assembly tasks are performed manually over multiple days, implementing smart factory systems remains a challenge. Previous work has proposed Anomaly Triplet-Net, which estimates assembly progress by applying deep metric learning to the visual features of products. Nevertheless, when visual changes between consecutive tasks are subtle, misclassification often occurs. To address this issue, this paper proposes a robust system for estimating assembly progress, even in cases of occlusion or minimal visual change, using a small-scale dataset. Our method leverages a Quadruplet Loss-based learning approach for anomaly images and introduces a custom data loader that strategically selects training samples to enhance estimation accuracy. We evaluated our approach using a image datasets: captured during desktop PC assembly. The proposed Anomaly Quadruplet-Net outperformed existing methods on the dataset. Specifically, it improved the estimation accuracy by 1.3\% and reduced misclassification between adjacent tasks by 1.9\% in the desktop PC dataset and  demonstrating the effectiveness of the proposed method.
\end{abstract}

\keywords{First keyword \and Second keyword \and More}

\section{Introduction}
\subsection{Research Background}
In recent years, advances in AI technologies have accelerated the digital transformation of manufacturing plants, a trend known as the Smart Factory initiative \cite{EXsmartfactory, smartfactory}. These efforts play a crucial role in increasing profitability. In particular, understanding the operational speed within the factory can lead to more efficient workflows. For products assembled from multiple components, managing component inventory and monitoring assembly progress are essential for minimizing the cost of excess part disposal. Therefore, smart factory systems are being increasingly applied to production lines.

\begin{figure}[t]
  \centering
  \includegraphics[width=0.6\linewidth]{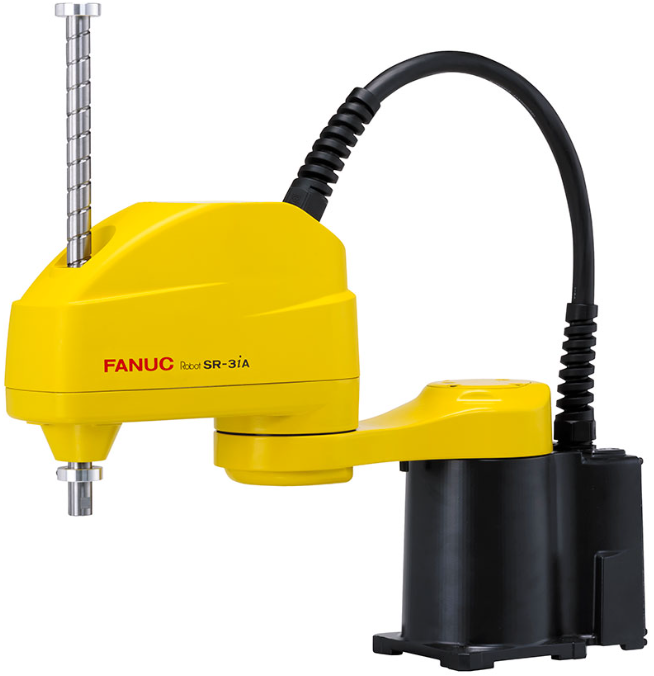}
  \caption{SCARA robot used in an assembly line \cite{robo1}}
  \label{Fig:robo1}
\end{figure}

\begin{figure}[t]
  \centering
  \includegraphics[width=0.6\linewidth]{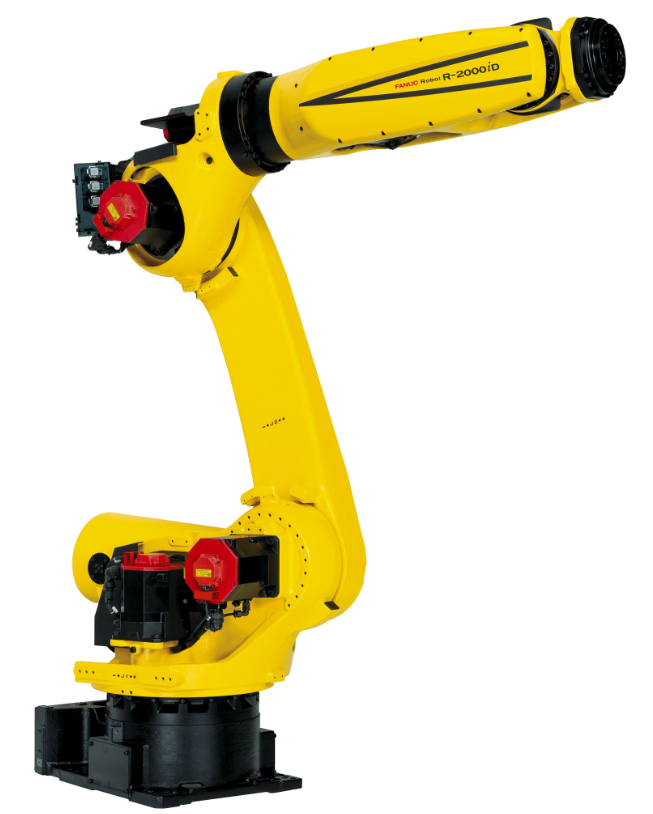}
  \caption{Large-scale welding robot used in an assembly line \cite{robo2}}
  \label{Fig:robo2}
\end{figure}

Examples of industrial robots used in such lines include SCARA robots (Fig. \ref{Fig:robo1}) and large-scale welding robots (Fig. \ref{Fig:robo2}). These robots can operate 24/7 and at speeds faster than humans, significantly improving production efficiency. However, while sensing technologies are relatively easy to deploy in products manufactured through a line production system, they face challenges in environments where products are assembled manually in irregular settings. A typical example of such environments is cell-based production systems.

\begin{figure}[t]
  \centering
  \includegraphics[width=0.9\linewidth]{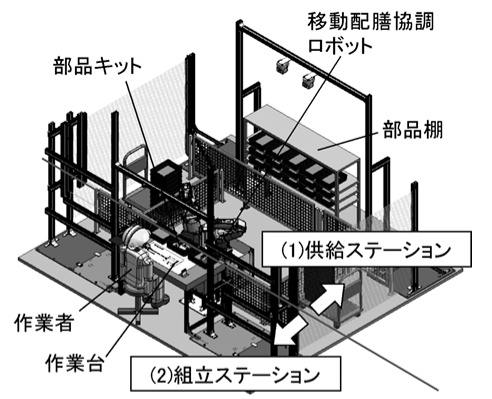}
  \caption{Overview of a human--robot collaborative cell production system \cite{cell1}}
  \label{Fig:cell1}
\end{figure}

\begin{figure}[t]
  \centering
  \includegraphics[width=0.9\linewidth]{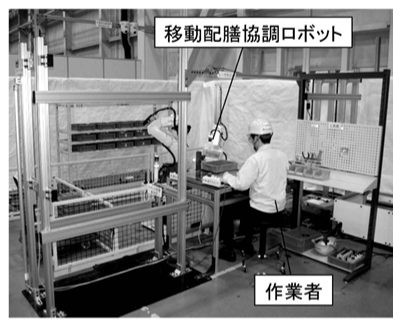}
  \caption{Example of actual collaborative work between a human and a robot \cite{cell1}}
  \label{Fig:cell2}
\end{figure}

Figure \ref{Fig:cell1} shows an overview of a collaborative cell production system with humans and robots, and Fig. \ref{Fig:cell2} depicts a human and robot working together. There are several reasons why sensing is difficult in such systems. First, since human assembly is often not strictly standardized, the procedures may vary from worker to worker. Second, in tasks that span several days, standardizing the procedure itself becomes challenging, and even if it is standardized, it may be difficult for workers to consistently follow such procedures.

As a result, smart factory practices have not been widely adopted in many real-world factories. Progress tracking is still often done manually, for example, by writing progress on paper and attaching it to the product. To address this, several studies have attempted to estimate progress through human motion recognition \cite{natu, Funk, Paula}. However, such approaches are often limited to relatively simple tasks that can be completed within short timeframes and may not be suitable for real-world factory environments.

To overcome these limitations, some research has focused on estimating progress based on the visual features of the product rather than human motion. One such study employs instance segmentation to recognize parts of the product, using this information to estimate assembly progress \cite{yumoto}. Their method is designed to achieve high accuracy even with a small dataset by masking the training images of individual parts, which helps to minimize the cost of real-world implementation.

However, in actual factory environments where occlusion frequently occurs, small parts can be completely hidden, reducing the accuracy of part detection. This in turn degrades the accuracy of progress estimation.

In contrast, another line of research focuses on capturing the overall appearance of the product rather than detecting individual parts \cite{kitsukawa}. This approach utilizes deep metric learning to estimate assembly progress. By treating images representing different assembly stages as training data, the task of progress estimation is reframed as a classification problem. The model is also trained using synthetically generated occluded images, enabling it to handle occlusion in practice. Experiments have shown successful progress estimation even in occluded scenarios.

However, a significant challenge remains: when consecutive tasks exhibit only slight differences in appearance, the model tends to misclassify them due to the small visual changes between stages.

To address this, the present study proposes a robust progress estimation system that focuses on the appearance features of assembled products. Specifically, we aim to build a system that can accurately estimate progress even under occlusion and when the visual changes between assembly steps are minimal, using training on a small-scale dataset.

\subsection{Related Work}
\label{sec:intro:related}

\subsubsection{Approaches Based on Skeletal Keypoint Information}
\label{sub:intro:hitachi}

Hitachi Industrial Control Solutions has proposed a method for estimating task progress by capturing skeletal keypoint information of workers as time-series data \cite{hitachi}. 
In this method, general-purpose cameras such as webcams or surveillance cameras are installed as fixed-position cameras to capture the worker's actions. From the recorded video, skeletal keypoints are extracted, and the system estimates task completion times based on movements and postures of these points.

One of the advantages of this approach is that it does not rely on wearable devices, thereby avoiding interference with the worker's movements. However, this method is sensitive to variations in physical build and working speed among individual workers, which may lead to inaccurate progress estimation. Additionally, in scenarios where product assembly spans several days, each short sub-task is often not strictly standardized and is instead performed in a personalized, ad-hoc manner. As a result, skeletal keypoint information alone may be insufficient for accurate progress estimation in such cases.\\

\subsubsection{Approaches Focusing on Product State}
\label{sub:intro:toshiba}

\begin{figure}[htbp]
  \centering
  \includegraphics[scale=1]{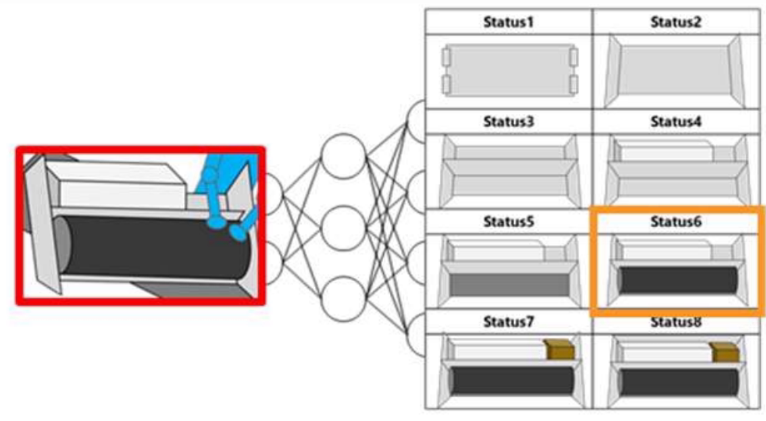}
  \caption{Progress estimation based on product state~\cite{toshiba}}
  \label{Fig:toshiba}
\end{figure}

Oshima et al. proposed a progress estimation method based on deep learning~\cite{toshiba}. An overview of the method is shown in Fig.~\ref{Fig:toshiba}. The method involves training a model using cropped product images that correspond to predefined assembly progress stages. These labeled images are input into ResNet~\cite{ResNet}, a type of Convolutional Neural Network (CNN), which outputs one of the predefined progress stages.

However, a common characteristic in many assembly processes is that the product's visual appearance does not significantly change until the assembly is nearly complete. This lack of visual distinction between stages may lead to misclassification, which poses a challenge for this approach.\\

\subsubsection{Approaches Focusing on Product Components}
\label{sub:intro:yumoto}

Yumoto et al. proposed a progress estimation method based on instance segmentation~\cite{yumoto}. 
The system first trains a YOLACT~\cite{YOLACT} model to detect the components required for product assembly. Then, by identifying the types of components present in a product, it estimates the current assembly stage.

The relationship between detected component combinations and progress stages is predefined. 
To reduce the cost of deployment in actual factories, data augmentation is performed using Random Erasing~\cite{RandomErasing}, which allows the model to be trained with a smaller dataset. This augmentation technique improves both component detection and progress estimation

However, this approach faces challenges in situations where components are small or distant from the camera. In such cases, occlusion may cause components to be entirely hidden, leading to failures in detection and, consequently, inaccurate progress estimation.\\

\subsection{Research Objective}
\label{sec:intro:purpose}
This study is conducted as a collaborative research project with an industrial partner. Therefore, the ultimate goal is to establish a method that can be deployed in real-world industrial settings. Currently, the progress of product assembly is manually recorded by workers on-site and managed by attaching handwritten labels to the products. 

However, from the factory's perspective, there is a strong demand to monitor the assembly progress remotely, using only fixed-position cameras without the need for direct site visits. Additionally, since on-site workers are generally not familiar with deep learning, it is essential to design a system that does not impose barriers such as the need to manually create datasets.

The experiments in this study are conducted using a dataset that simulates factory products, due to intellectual property and confidentiality constraints that prevent the direct use of actual factory product data. 
Although a simulated dataset is employed for experimentation, the intended target of the proposed method is real factory products.

\begin{figure}[htbp]
  \centering
  \includegraphics[scale=1.25]{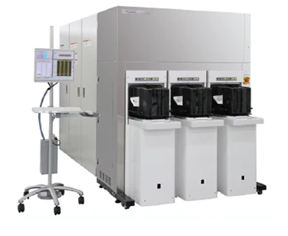}
  \caption{High-Acceleration SEM Measurement System~\cite{oogatasouchi}}
  \label{Fig:oogatasouchi}
\end{figure}

The target product considered in this study is a semiconductor inspection device, which typically requires three to four days to assemble. 
An example of such a device is shown in Fig.~\ref{Fig:oogatasouchi}.

\begin{figure}[htbp]
  \centering
  \includegraphics[scale=0.5]{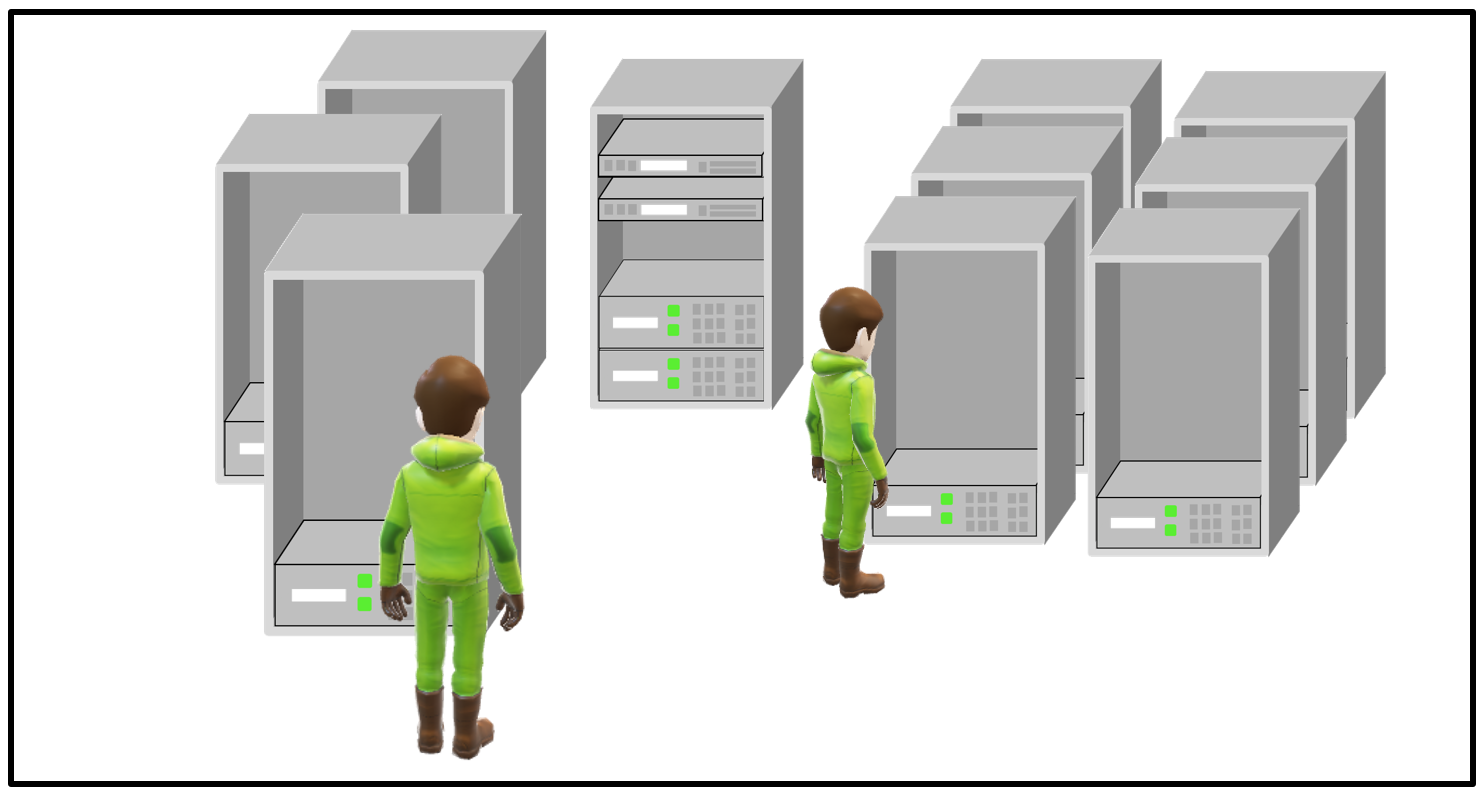}
  \caption{Example of an Assumed Factory Environment}
  \label{Fig:factory_image}
\end{figure}

\begin{figure}[htbp]
  \centering
  \includegraphics[scale=0.25]{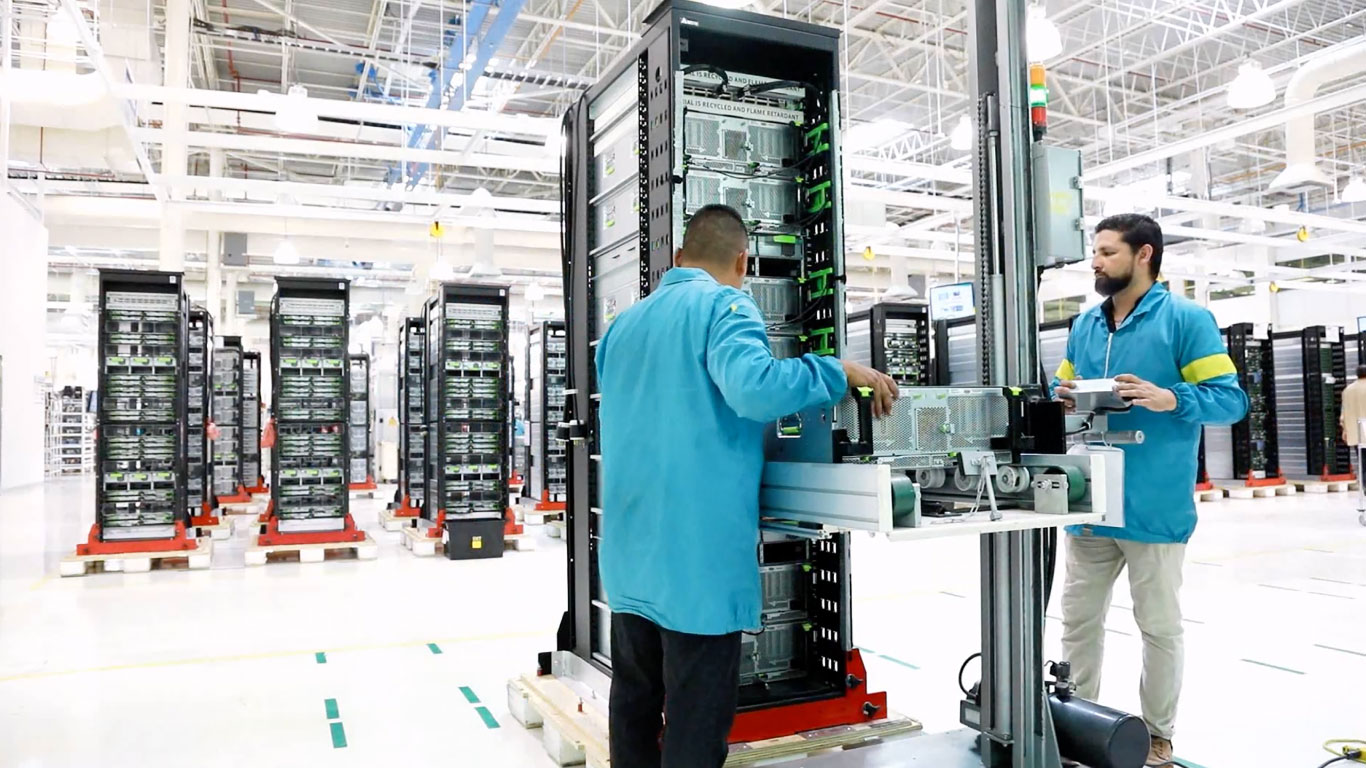}
  \caption{Example of Large-Scale Server Assembly (1)~\cite{server_1}}
  \label{Fig:server_1}
\end{figure}

\begin{figure}[htbp]
  \centering
  \includegraphics[scale=1.3]{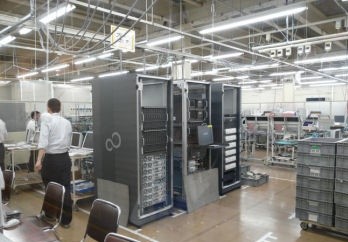}
  \caption{Example of Large-Scale Server Assembly (2)~\cite{server_2}}
  \label{Fig:server_2}
\end{figure}

Due to intellectual property constraints, we cannot include actual factory images from the collaborating company in this paper. Instead, we provide an illustrative image of a factory environment in Fig.~\ref{Fig:factory_image}. In addition, we present examples of large-scale assembly tasks, such as server assembly, in Fig.~\ref{Fig:server_1} and Fig.~\ref{Fig:server_2}.

Camera placement is subject to physical limitations, and the captured product videos often suffer from occlusions caused by other products, workers, or tools. Moreover, because the camera is positioned at a distance, the external appearance of the assembled product may change only slightly, making it difficult to perceive these changes without domain knowledge.

Therefore, the objective of this research is to develop a robust system that can operate on small-scale datasets and focus on the visual features of the assembly target, while being resilient to occlusion and subtle appearance changes during the assembly process.

\clearpage
\section{Proposedmethod}
\label{sec:Proposedmethod}
\subsection{Assembly Progress Estimation Method}
\label{sec:ProposedMethod:AssemblyProgressEstimationMethod}
\begin{figure}[htbp]
  \centering
  \includegraphics[scale=0.5]{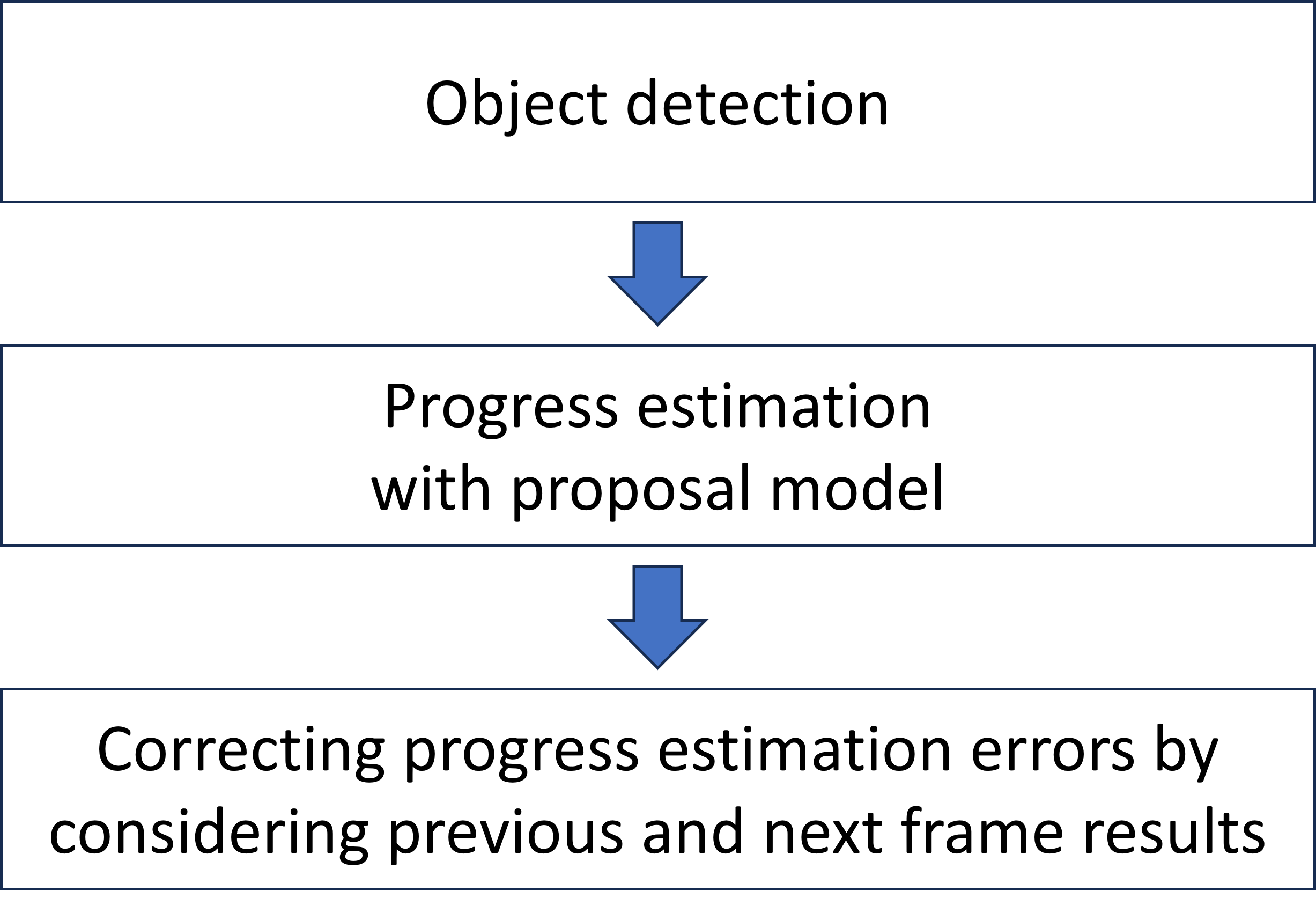}
  \caption{Diagram of the proposed system}
  \label{Fig:ProposedMethod}
\end{figure}

An overview of the proposed system is shown in Fig. \ref{Fig:ProposedMethod}. This study builds upon the Anomaly Triplet-Net framework, introducing several improvements to achieve higher progress estimation accuracy. The key enhancement lies in replacing the original progress estimation model with a newly proposed model that processes detected product images. Additionally, several modifications have been implemented, including a redesign of the data loader used during training, adjustments to the dimensionality of the feature space, and an improved loss function.

The proposed method particularly focuses on reducing misclassifications between adjacent tasks—those that involve subtle changes in visual appearance and are performed consecutively. By addressing this issue, the method is expected to improve the accuracy of progress estimation when applied to video data with temporal processing. Specifically, in previous studies, temporal modeling failed to improve accuracy in scenarios involving such adjacent tasks. The proposed improvements aim to overcome this limitation and enhance estimation performance. 

\subsection{Assembly product etection}
\label{sec:ProposedMethod:Assemblyproductdetection}
The target objects for progress estimation are detected using the object detection algorithm YOLOv8\cite{YOLOv8}. A custom dataset is created, and fine-tuning is performed. Fine-tuning refers to retraining the weights of a pre-trained network starting from its pre-initialized weights.

In this study, considering the possibility that factory workers with limited AI knowledge may be responsible for dataset creation, it is desirable to minimize the dataset creation cost. Therefore, the custom dataset for YOLOv8 and the dataset for the proposed Anomaly Quadruplet-Net are constructed simultaneously.

Initially, a small number of YOLO-format annotation samples are created. A model is trained using this small dataset, and used to crop the product images during inference. Subsequently, the correctly detected samples are visually confirmed and categorized into progress stages (Steps).

The dataset divided by Steps is used for the Anomaly Quadruplet-Net, while the unified dataset without Step categorization is used for YOLO training. During inference, YOLO trained on the custom dataset detects product regions, which are then cropped and passed to the deep metric learning model for progress estimation.

 \subsection{Anomaly Quadruplet-net:modelsthatconsider occlusion}
\label{sec:AnomalyQuadruplet}

\subsubsection{Overall Structure of Anomaly Quadruplet-Net}
\label{sec:AnomalyQuadruplet:Overall}
\begin{figure}[h]
    \centering
    \includegraphics[width=1\linewidth]{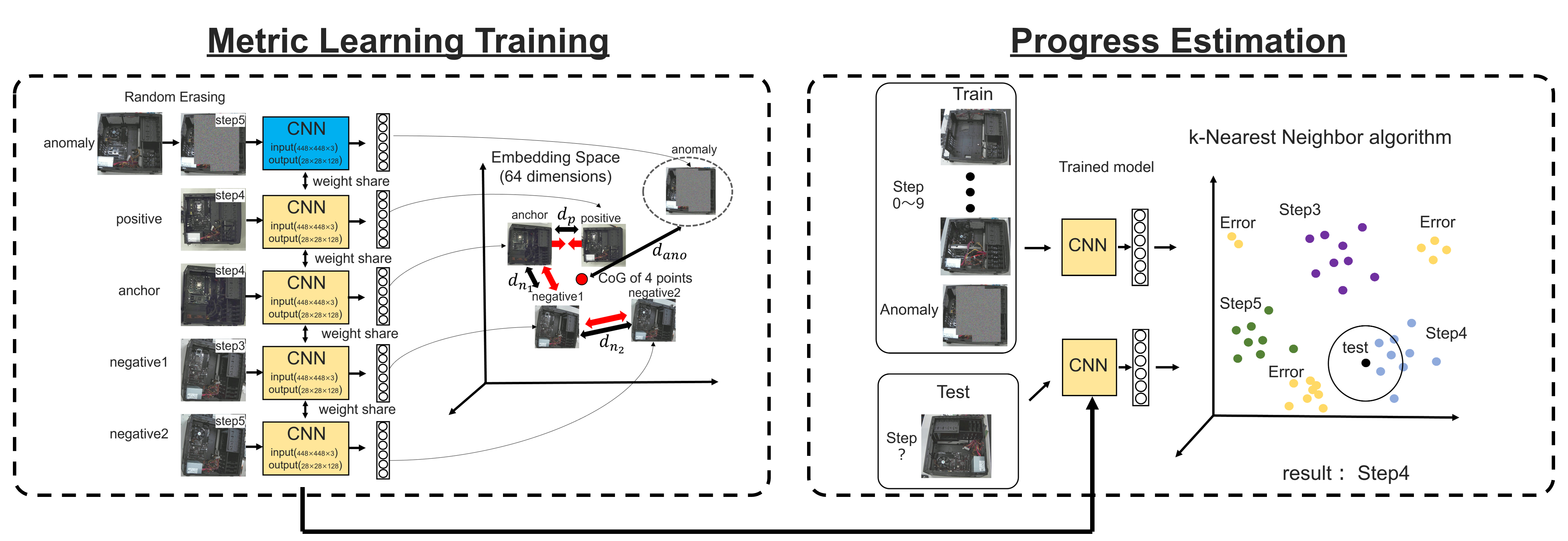}
    \caption{Overall structure of Anomaly Quadruplet-Net}
    \label{Fig:OverallAnomalyQuadruplet}
\end{figure}

The overall architecture of the proposed progress estimation model, Anomaly Quadruplet-Net, is shown in Fig.~\ref{Fig:OverallAnomalyQuadruplet}. The model consists of two phases: a training phase and an inference phase. In the training phase, the weights of the CNN are optimized. In the inference phase, the learned weights are used to map unseen data into a feature space.

To handle classes with subtle visual differences, a custom data loader is designed that strategically selects training samples for the Quadruplet Loss\cite{QuadrupletLoss}. \\

\subsubsection{Training Phase}
\label{sec:AnomalyQuadruplet:Train}

\begin{figure}[h]
    \centering
    \includegraphics[width=1\linewidth]{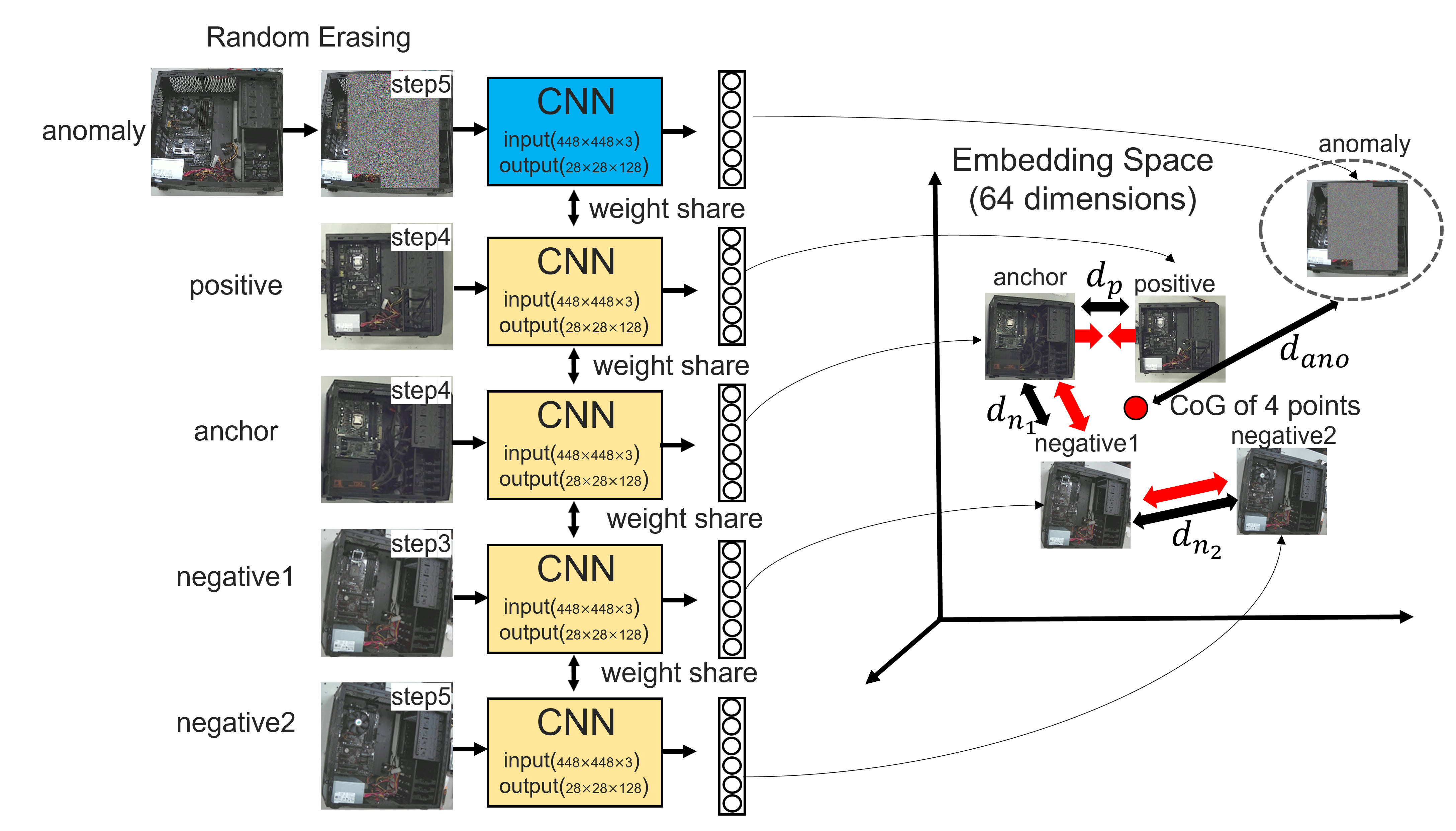}
    \caption{Model architecture of the training phase}
    \label{Fig:AnomalyQuadrupletTrain}
\end{figure}

The model architecture for the training phase is shown in Fig.~\ref{Fig:AnomalyQuadrupletTrain}. First, progress steps are defined based on the assembly procedure, and training is prepared accordingly. From the assembly videos, objects for progress estimation are cropped and saved for each corresponding step.

Using the saved step images, five types of training samples are prepared: anchor, positive, negative1, negative2, and anomaly. These samples are then used for training. Each sample is passed through a CNN feature extractor. The CNN consists of four convolutional layers and takes a 448×448×3 image (height 448, width 448, 3 channels) as input, outputting a 28×28×128 feature map. The weights of the CNN are shared across all inputs.

Following the CNN layers, the features are passed through a fully connected layer to project them into a feature space of arbitrary dimension. Distance metric learning is then performed in this feature space using the proposed loss function.

\begin{figure}[h]
    \centering
    \includegraphics[width=1\linewidth]{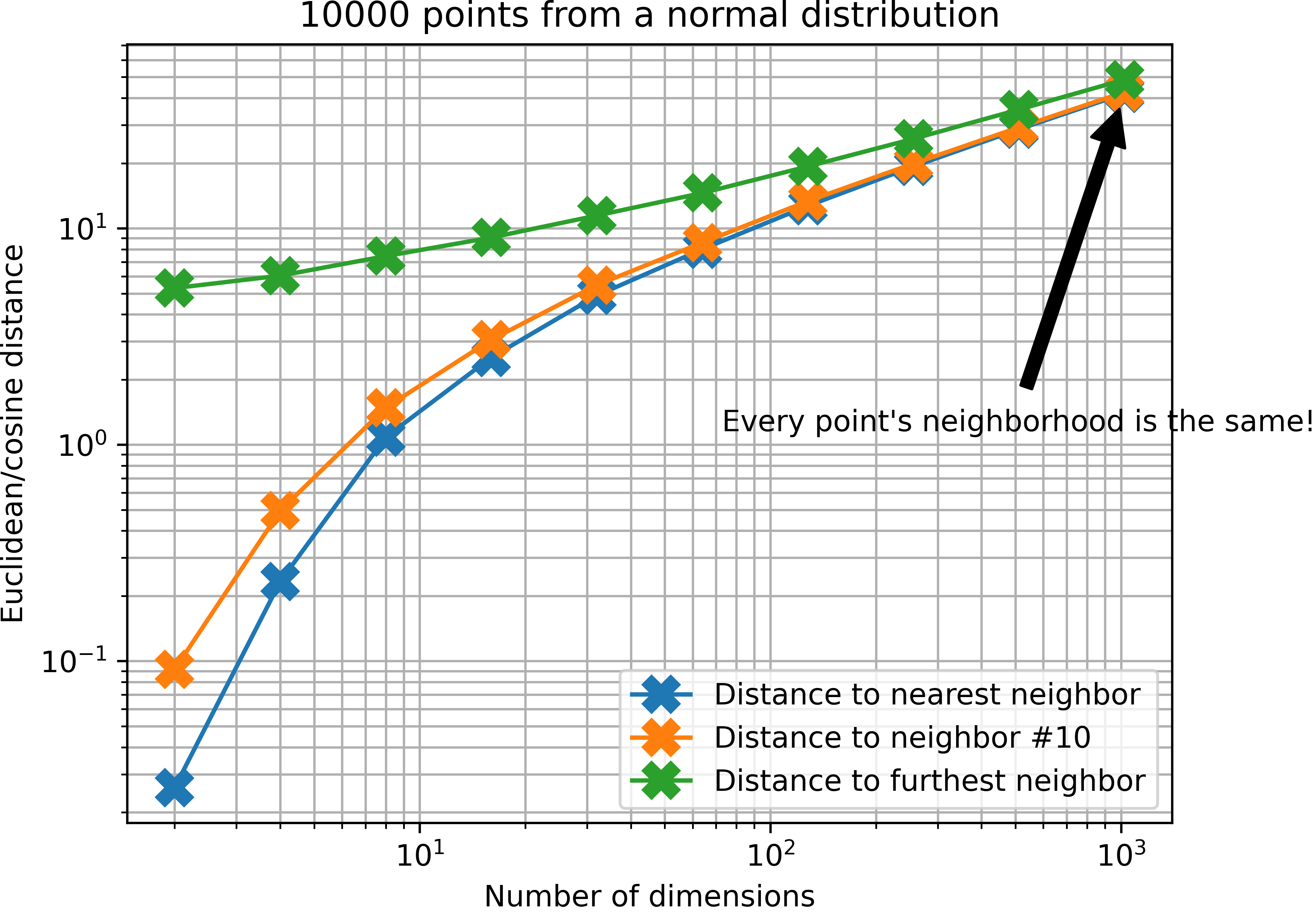}
    \caption{Illustration of the Curse of Dimensionality \cite{CurseOfDimensionality_Web}}
    \label{Fig:CurseOfDimensionality_Web}
\end{figure}

Anomaly Quadruplet-Net performs metric learning by embedding features extracted by a CNN into a feature space with an arbitrary number of dimensions. Therefore, it is necessary to determine the dimensionality of this feature space. The appropriate dimensionality depends on factors such as the patterns of input images and the number of training epochs, and is typically decided based on empirical knowledge and experimental results. However, when the dimensionality exceeds two digits, it is possible that distances in the feature space become almost indistinguishable. To avoid this phenomenon, known as the "curse of dimensionality," it is considered necessary to limit the dimensionality to two digits or fewer. Fig.~\ref{Fig:CurseOfDimensionality_Web}~\cite{CurseOfDimensionality} illustrates the curse of dimensionality.

\begin{figure}[h]
    \centering
    \includegraphics[width=1\linewidth]{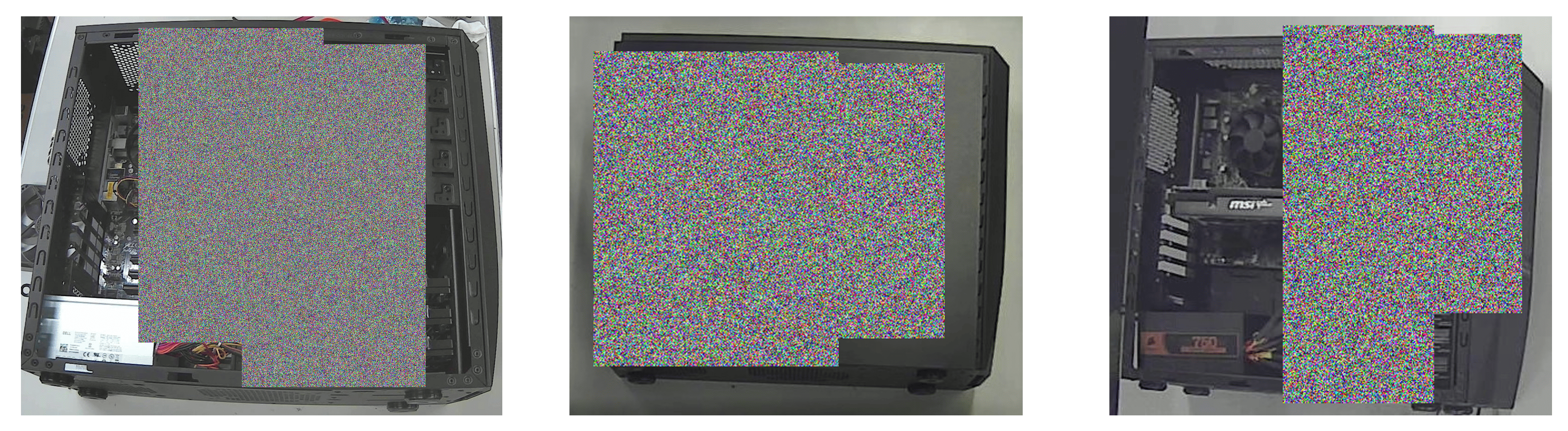}
    \caption{Anomaly samples generated by adding noise using Random Erasing}
    \label{Fig:RandomErasing}
\end{figure}

The generation method of anomaly samples utilizes Random Erasing \cite{RandomErasing} to mask a randomly selected rectangular region of an image with random pixel values. As an example, an image with noise applied by Random Erasing is shown in Fig. \ref{Fig:RandomErasing}. By obscuring part of the product image, we create pseudo-occlusion images that even humans cannot use to estimate the assembly progress.

Additionally, the settings for Random Erasing used in this method are described below.
During training, Random Erasing is applied only to input images selected from classes other than the class to which the anomaly sample belongs.
The sample selection procedure during training is designed so that images from non-anomalous classes are used as targets for Random Erasing, thereby preventing the degradation of anomaly-specific features.

The ratio of the area of the rectangular region to the input image area is randomly selected in the range from 0.02 to 0.4.
The aspect ratio of the rectangular region is randomly chosen between 0.3 and 3.3.
This Random Erasing operation is applied twice to each image.

Regarding the loss function, it was designed taking into consideration tasks where occlusion and changes in appearance features are small.We improve this method by improving the positional relationship of each sample in the conventional method, Anomaly Triplet-Net, and incorporating Quadruplet Loss.
The proposed loss function is shown in equation (\ref{eq:AnomalyQuadruplet}).

\begin{equation}
L_{\mathrm{AnomalyQuadruplet}}
=
\max(d_p - d_{n1} + m_\alpha, 0)
+
\max(d_p - d_{n2} + m_\beta, 0)
+
\lambda \,
\max(d_p - d_{ano} + m_c, 0)
\label{eq:AnomalyQuadruplet}
\end{equation}

$d_p$, $d_{n1}$, and $d_{n2}$ represent the distances in the feature space between the anchor sample and the positive sample, negative1 sample, and negative2 sample, respectively.
$d_{ano}$ denotes the distance in the feature space between the anomaly sample and the centroid of the four samples: the anchor, positive, negative1, and negative2 samples.

$m_\alpha$, $m_\beta$, and $m_c$ are margin parameters that define the target distances between samples during training.
$\lambda$ is a weighting coefficient that adjusts the contribution of the term involving $d_p$ and $d_{ano}$ relative to the terms involving $d_p$ and $d_{n1}$, and $d_p$ and $d_{n2}$. In other words, $\lambda$ controls the epoch at which training with anomaly samples begins.

With this design, the model first focuses on sufficiently learning the Quadruplet Loss component. After the Quadruplet Loss has been adequately learned, training using anomaly samples is gradually introduced, enabling more effective learning of anomaly samples.

As an example, consider a training process consisting of 100 epochs, where learning with anomaly samples starts from the 50th epoch. In this case, $\lambda$ is set to 0 until the 49th epoch, and then gradually increases from 0 to 1 between the 50th and 100th epochs.

\begin{figure}[h]
    \centering
    \includegraphics[width=1\linewidth]{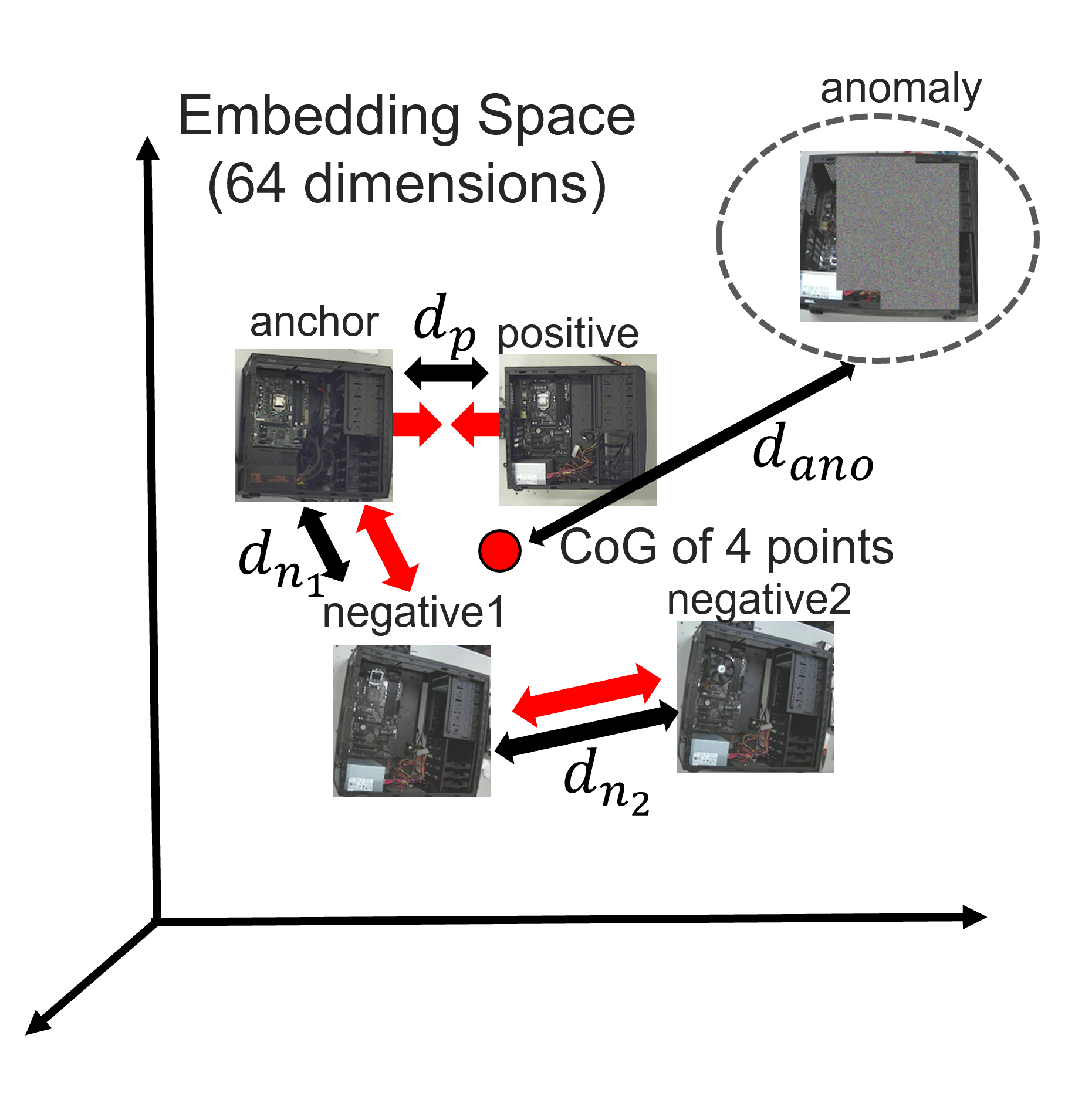}
    \caption{Relative positions of each sample in the feature space}
    \label{Fig:AnomalyQuadruplet_Loss}
\end{figure}

To enable recognition even when changes in appearance features are small, and to minimize computational cost, the data loader is carefully designed. By constructing the dataset appropriately, misclassification between adjacent assembly steps can be reduced, while also improving the required PC specifications and processing time for training.

When a dataset contains $n$ classes and each class consists of $m$ images, the number of possible data combinations for Triplet Loss\cite{TripletLoss} is expressed by Eq.~(\ref{eq:anotriplet_c}), and that for Quadruplet Loss is expressed by Eq.~(\ref{eq:anoquadru_c}).

\begin{equation}
    {}_nC_1 \times {}_mP_2 \times {}_{n-1}C_1 \times {}_mP_1
    \label{eq:anotriplet_c}
\end{equation}

\begin{equation}
    {}_nC_1 \times {}_mP_2 \times {}_{n-1}C_1 \times {}_mP_1 \times {}_{n-2}C_1 \times {}_mP_1
    \label{eq:anoquadru_c}
\end{equation}

As an example, when the dataset consists of eight classes with forty images per class, the number of combinations is approximately 3.5 million for Triplet Loss and approximately 800 million for Quadruplet Loss. Such a massive computational cost poses a serious problem in terms of processing time and computational resources.

To address this issue, the proposed method focuses on the fact that consecutive classes correspond to consecutive assembly steps and therefore exhibit similar appearance features. By selecting only combinations among three consecutive classes, the number of training data combinations is significantly reduced.

In Anomaly Quadruplet-Net, one anchor sample is selected from each of the $n$ classes. A positive sample is selected from the same class as the anchor but from a different image. The negative1 and negative2 samples are selected from the classes immediately preceding and succeeding the anchor class, respectively. The anomaly sample is generated by applying Random Erasing \cite{RandomErasing} to an image selected from a class other than the anchor class.

To sufficiently separate similar classes in the feature space, the classes of the negative1 and negative2 samples are swapped and the combinations are generated again. As a result, the total number of combinations is reduced to $2 \times n \times m$. This design preserves the cluster structure across all classes and enables correct learning of the global structure of the feature space.

\begin{figure}[h]
    \centering
    \includegraphics[width=1\linewidth]{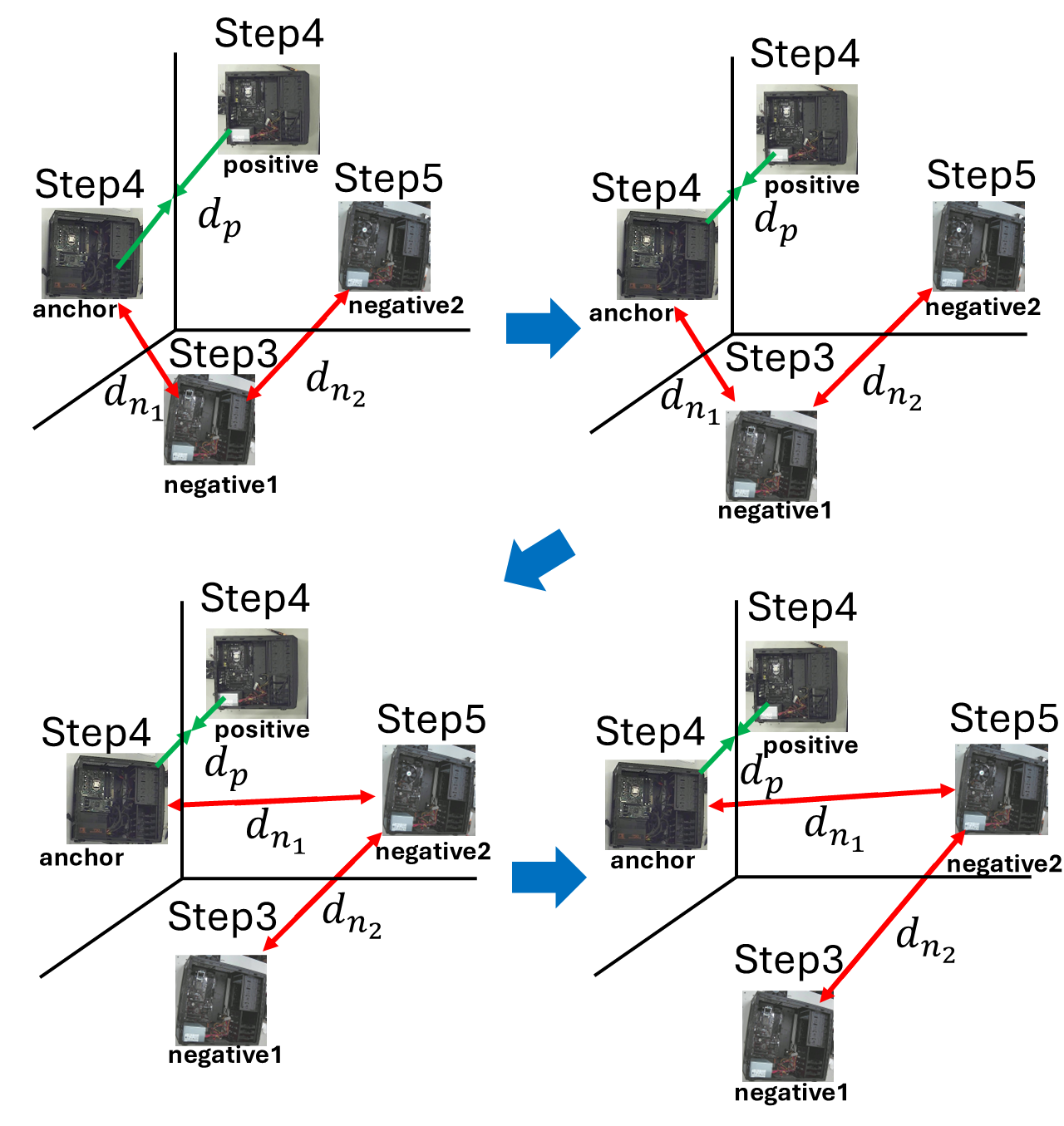}
    \caption{Sample combinations used during training}
    \label{Fig:QuadrupletDataLoader}
\end{figure}

Fig.\ref{Fig:QuadrupletDataLoader} illustrates the training data combination strategy.

\clearpage
\section{Inference Phase}
\label{sec:AnomalyQuadruplet:Estimate}

During the inference phase, a trained model obtained in the training phase is used. For unknown input data, estimation is performed by searching for nearby samples in the feature space among the training data. In this manner, the method identifies the closest class from both the classes included in the training dataset and anomaly images that are not included in the training data, and outputs the result as the progress estimation.

\subsection{Step Estimation Using the k-Nearest Neighbor Method}
\label{sub:AnomalyQuadruplet:kNN}

\begin{figure}[h]
    \centering
    \includegraphics[width=1\linewidth]{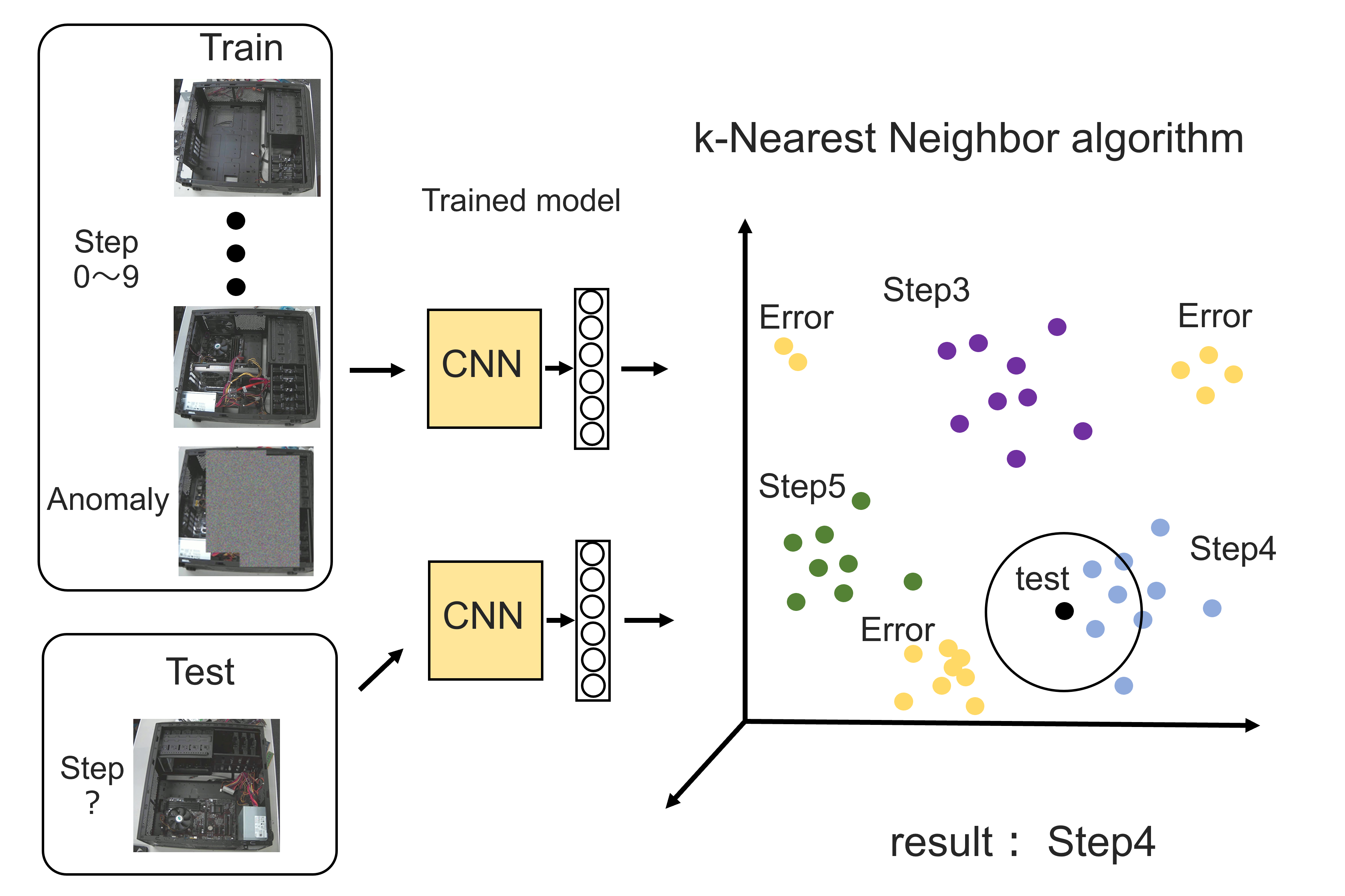}
    \caption{Model structure in the inference phase}
    \label{Fig:kNN}
\end{figure}

The inference model is illustrated in Fig.~\ref{Fig:kNN}. Similar to the previous work, Anomaly Triplet-Net, images from each step used during training are first individually input into the trained model and embedded into the feature space, where they are converted into feature vectors. Subsequently, an input image to be estimated is embedded into the same feature space using the trained model.  
By applying the k-nearest neighbor (kNN) algorithm \cite{kNN} to these embedded data, the step to which the unknown input data belongs is estimated.

In practical environments, occlusions may occur in product images due to dense arrangements of products or tools, or while a human operator is performing assembly tasks. Collecting such conditions in advance as a dataset and learning them explicitly is difficult, especially in industrial applications.  
Therefore, in the proposed method, pseudo-occluded images are generated as anomaly samples and embedded into the feature space during training.

However, when estimating unknown error images, it is possible that their embeddings are located far from those of the trained pseudo-occlusion images in the feature space. To handle such cases, a distance threshold is introduced in addition to the progress estimation based on the kNN method. If the distance between an unknown sample and the training samples exceeds a predefined threshold, the kNN-based estimation is not applied.

\begin{figure}[h]
    \centering
    \includegraphics[width=1\linewidth]{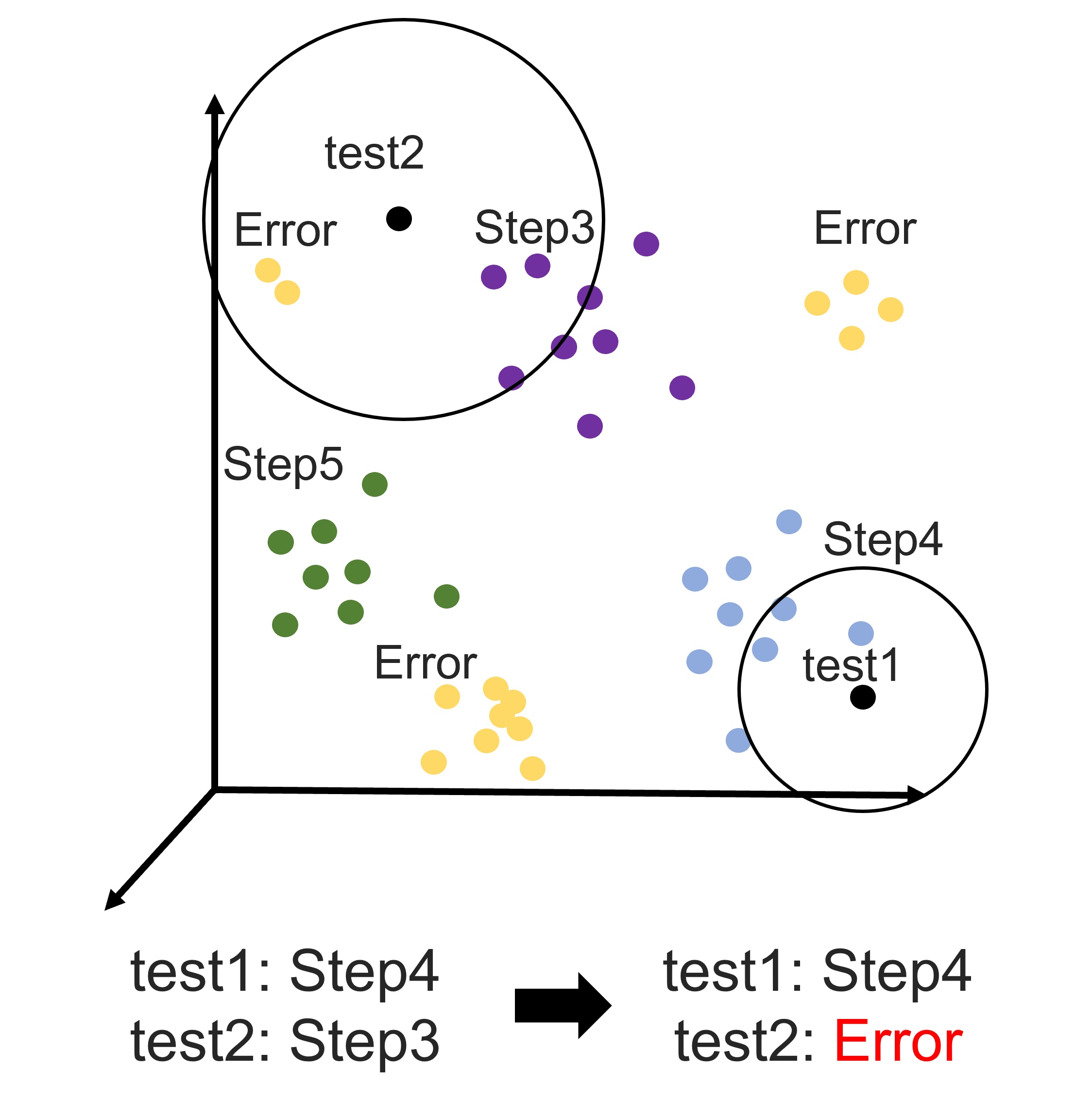}
    \caption{Error determination using a distance threshold in the feature space}
    \label{Fig:kNNthreshold}
\end{figure}

Fig.\ref{Fig:kNNthreshold} illustrates the error determination mechanism. According to the kNN-based estimation, test1 and test2 are classified as belonging to Step4 and Step3, respectively. However, since test2 is located far from Step3 in the feature space, it is corrected to an error judgment by applying the distance threshold.  
The distance used for this judgment is calculated as the average distance between the test sample and its $k$ nearest neighbors.

\clearpage
\subsection{Experiments}
This subsection describes the evaluation experiments of the proposed Anomaly Quadruplet-Net. 
The evaluation was conducted by comparing the proposed Anomaly Quadruplet-Net with a previous method whose parameters were refined. 
In order to verify the effectiveness of the proposed Anomaly Quadruplet-Net, comparative experiments were performed with the previous method, Anomaly Triplet-Net. 
For this purpose, experiments were conducted on the training parameters of Anomaly Triplet-Net. 
Specifically, experiments were carried out while changing the total number of training epochs, the epoch at which anomaly learning starts, the dimensionality of the feature space of the model, the value of $k$ in kNN, and the error determination threshold during inference. 
In this subsection, effective training parameters for the previous method are discussed, and by obtaining progress estimation results using the derived parameters, they are used for comparison experiments with the proposed Anomaly Quadruplet-Net. 
Desktop PC assembly images were used in the experiments.

\subsubsection{Experimental Environment}
\begin{figure}[h]
    \centering
    \includegraphics[width=1\linewidth]{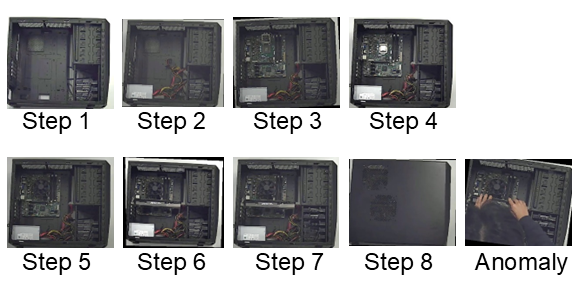}
    \caption{Desktop PC dataset}
    \label{Fig:DesttopPCdataset}
\end{figure}

First, the same dataset as that used in the previous study, Anomaly Triplet-Net, was used for the experiments. 
Therefore, a desktop PC dataset was employed. 
Figure~\ref{Fig:DesttopPCdataset} shows each step of the desktop PC assembly images. 
The images were created by recording the desktop PC assembly process as a video using a camera and then extracting product images from the video to construct the dataset. 
As the camera, a network camera Qwatch TS-WRLP manufactured by I-O DATA DEVICE, Inc.\cite{Qwatch} was used, and the image resolution was set to $1980 \times 1080$.

\subsubsection{Experimental Conditions}
The steps to be classified consist of eight steps from Step~1 to Step~8, in addition to an Error class corresponding to anomaly detection, resulting in a total of nine classes. 
The criteria for each step are described as follows. 
Step~1 represents the state in which only the PC case is present. 
Step~2 represents the state in which the power supply unit is installed. 
Step~3 represents the state in which the motherboard is installed. 
Step~4 represents the state in which the CPU is installed on the motherboard. 
Step~5 represents the state in which a CPU cooler is installed on the CPU mounted on the motherboard. 
Step~6 represents the state in which a GPU is installed on the motherboard. 
Step~7 represents the state in which an HDD is installed in the PC case. 
Step~8 represents the state in which the PC case cover is attached. 
As anomaly data, only states in which occlusion caused by a human occurs were added to the dataset. 
It should be noted that anomaly data are not included in the training and validation datasets and are not used during training. 
That is, anomaly data are used only when testing the trained model.

The desktop PC images constituting the dataset are the same as those used in the dataset described in the Anomaly Triplet-Net paper. 
The training dataset consists of a total of 320 images, with 40 images for each step from Step~1 to Step~8. 
The validation dataset consists of a total of 1,264 images, with 150 images for each step. 
The test dataset consists of a total of 1,600 images, including images from Step~1 to Step~8 as well as Error images.

\subsubsection{Experimental Results and Discussion}
\begin{figure}[h]
    \centering
    \includegraphics[width=0.7\linewidth]{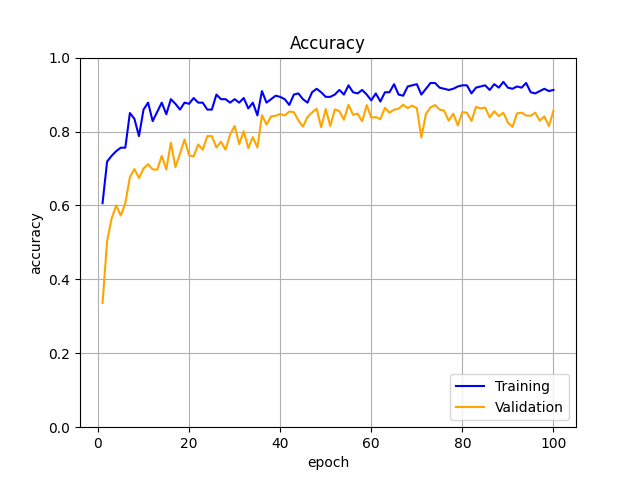}
    \caption{AnomalyTriplet Condition 1: Transition of accuracy with respect to the number of training epochs}
    \label{Fig:AnomalyTripletResult_1_Accuracy}
\end{figure}

\begin{figure}[h]
    \centering
    \includegraphics[width=0.7\linewidth]{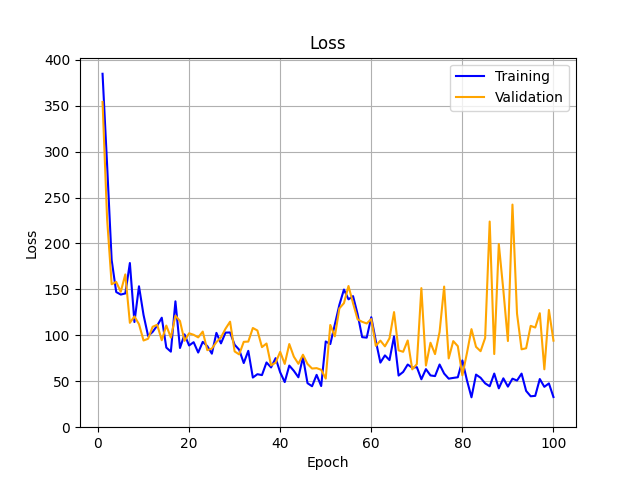}
    \caption{AnomalyTriplet Condition 1: Transition of loss with respect to the number of training epochs}
    \label{Fig:AnomalyTripletResult_1_Loss}
\end{figure}

\begin{figure}[h]
    \centering
    \includegraphics[width=0.7\linewidth]{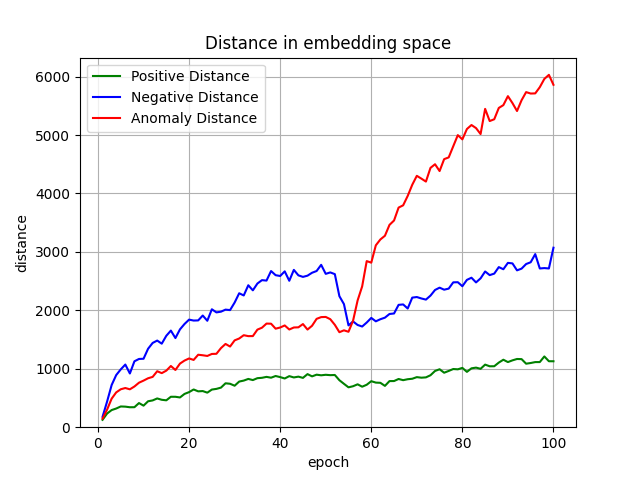}
    \caption{AnomalyTriplet Condition 1: Distances between samples in the feature space}
    \label{Fig:AnomalyTripletResult_1_Distance}
\end{figure}

\begin{figure}[h]
    \centering
    \includegraphics[width=0.7\linewidth]{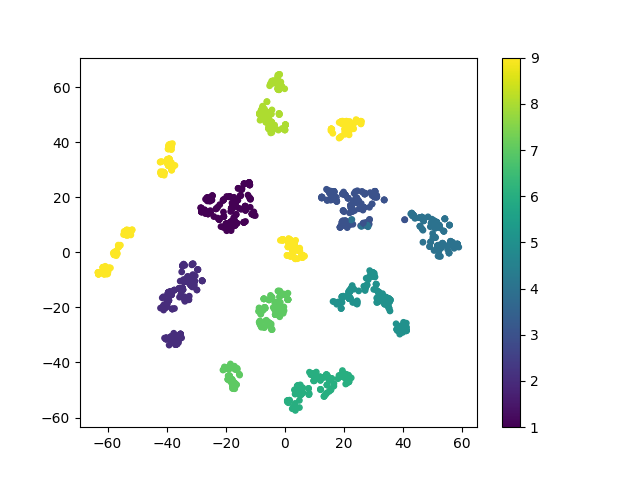}
    \caption{AnomalyTriplet Condition 1: Two-dimensional visualization of the feature space using t-SNE}
    \label{Fig:AnomalyTripletResult_1_T-SNE}
\end{figure}

\begin{figure}[h]
    \centering
    \includegraphics[width=0.7\linewidth]{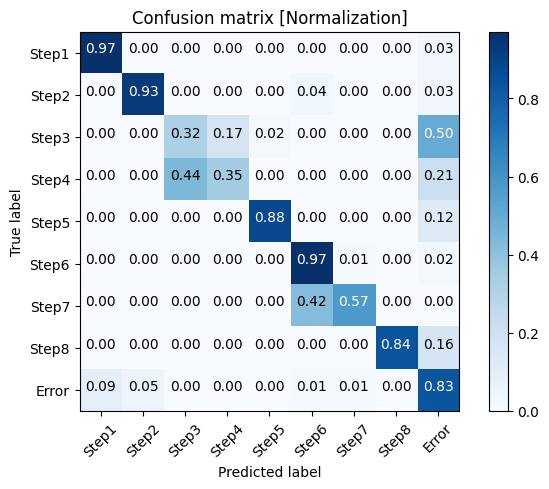}
    \caption{AnomalyTriplet Condition 1: Evaluation using a confusion matrix}
    \label{Fig:AnomalyTripletResult_1_normalize_test}
\end{figure}

First, experiments were conducted under conditions close to those described in the original Anomaly Triplet-Net paper. 
The training conditions were set as follows: the total number of epochs was 100, anomaly learning started from epoch 50, $k=10$ for kNN, the dimensionality of the feature space was 128, and the learning rate was set to $lr=0.0001$. 
This condition is referred to as \textit{AnomalyTriplet Condition 1}. 
The results of training, including accuracy, loss, distances in the feature space, and two-dimensional visualization of the feature space using t-SNE\cite{tSNE}, are shown in Fig.~\ref{Fig:AnomalyTripletResult_1_Accuracy}, Fig.~\ref{Fig:AnomalyTripletResult_1_Loss}, Fig.~\ref{Fig:AnomalyTripletResult_1_Distance}, and Fig.~\ref{Fig:AnomalyTripletResult_1_T-SNE}, respectively. 
In addition, the confusion matrix of the progress estimation results when the error determination distance threshold was set to 1,425,000 is shown in Fig.~\ref{Fig:AnomalyTripletResult_1_normalize_test}. 
The accuracy for the test dataset was 74.3\%.

From Fig.~\ref{Fig:AnomalyTripletResult_1_Accuracy} and Fig.~\ref{Fig:AnomalyTripletResult_1_Loss}, it can be observed that the estimation accuracy for the validation dataset decreases while the loss increases, indicating that overfitting occurs. 
Furthermore, Fig.~\ref{Fig:AnomalyTripletResult_1_normalize_test} shows that misclassification occurs between Step~3 and Step~4. 
On the other hand, Fig.~\ref{Fig:AnomalyTripletResult_1_Distance} and Fig.~\ref{Fig:AnomalyTripletResult_1_T-SNE} indicate that distance learning itself in the feature space is successfully performed.

\begin{figure}[h]
    \centering
    \includegraphics[width=0.7\linewidth]{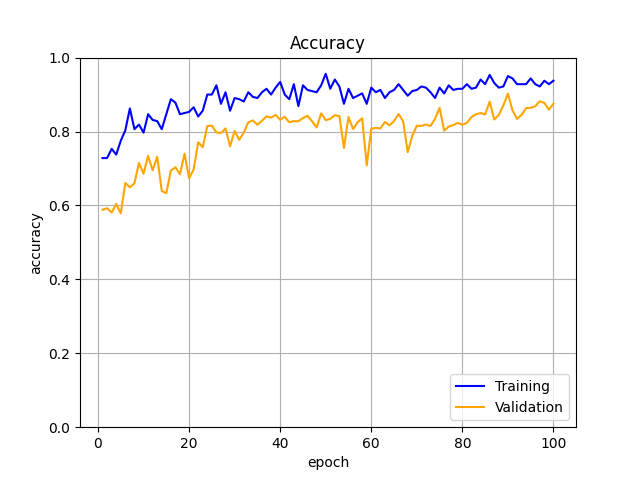}
    \caption{AnomalyTriplet Condition 2: Transition of accuracy with respect to the number of training epochs}
    \label{Fig:AnomalyTripletResult_2_Accuracy}
\end{figure}

\begin{figure}[h]
    \centering
    \includegraphics[width=0.7\linewidth]{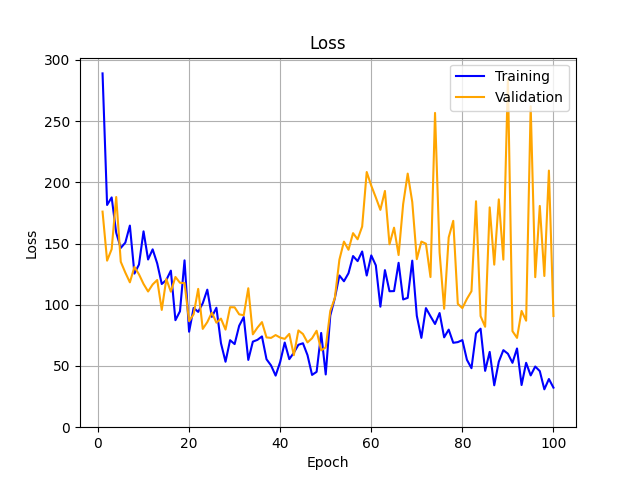}
    \caption{AnomalyTriplet Condition 2: Transition of loss with respect to the number of training epochs}
    \label{Fig:AnomalyTripletResult_2_Loss}
\end{figure}

\begin{figure}[h]
    \centering
    \includegraphics[width=0.7\linewidth]{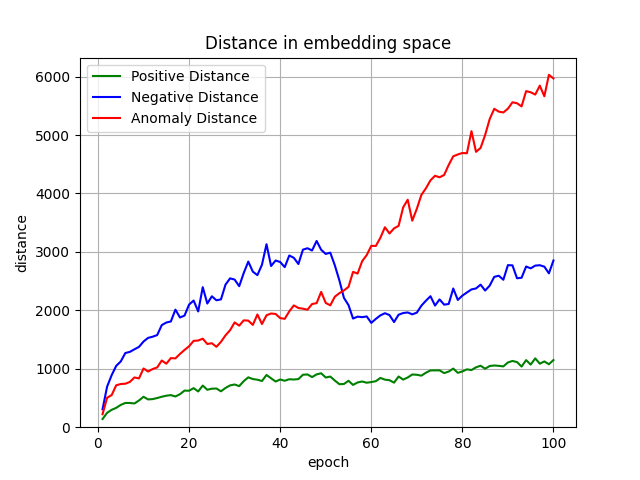}
    \caption{AnomalyTriplet Condition 2: Distances between samples in the feature space}
    \label{Fig:AnomalyTripletResult_2_Distance}
\end{figure}

\begin{figure}[h]
    \centering
    \includegraphics[width=0.7\linewidth]{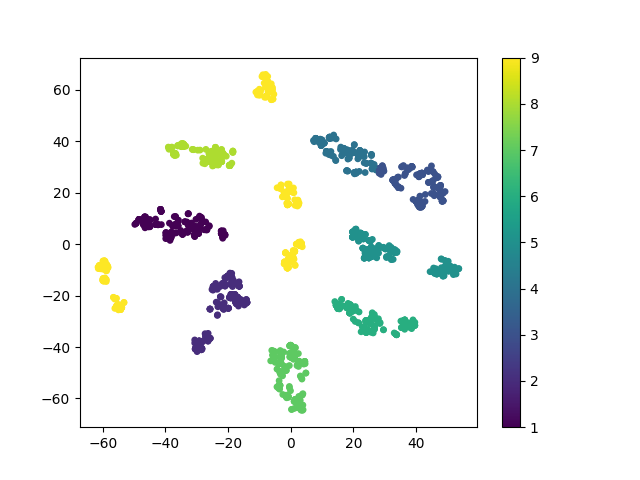}
    \caption{AnomalyTriplet Condition 2: Two-dimensional visualization of the feature space using t-SNE}
    \label{Fig:AnomalyTripletResult_2_T-SNE}
\end{figure}

\begin{figure}[h]
    \centering
    \includegraphics[width=0.7\linewidth]{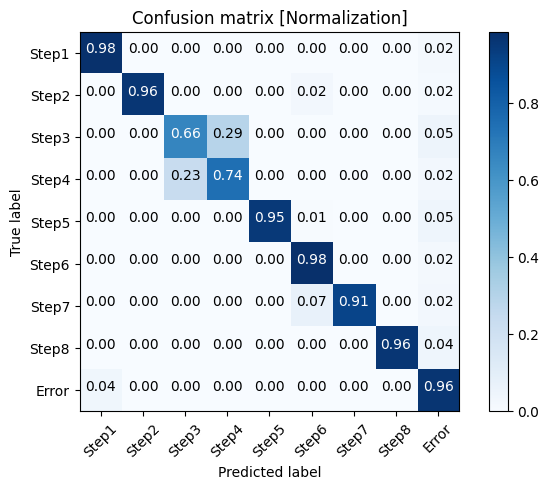}
    \caption{AnomalyTriplet Condition 2: Evaluation using a confusion matrix}
    \label{Fig:AnomalyTripletResult_2_normalize_test}
\end{figure}

\clearpage
Even though distances among samples in the feature space were learned, the feature space itself is a high-dimensional space with three-digit dimensionality, which may cause the curse of dimensionality. 
Therefore, the dimensionality of the feature space was reduced and experiments were conducted again. 
The training conditions were set such that the dimensionality of the feature space was reduced to 64, while all other conditions remained the same as in AnomalyTriplet Condition~1. 
This condition is referred to as \textit{AnomalyTriplet Condition 2}. 
The results of training, including accuracy, loss, distances in the feature space, and two-dimensional visualization using t-SNE, are shown in Fig.~\ref{Fig:AnomalyTripletResult_2_Accuracy}, Fig.~\ref{Fig:AnomalyTripletResult_2_Loss}, Fig.~\ref{Fig:AnomalyTripletResult_2_Distance}, and Fig.~\ref{Fig:AnomalyTripletResult_2_T-SNE}, respectively. 
As in Condition~1, the confusion matrix of the progress estimation results with the error determination distance threshold set to 1,425,000 is shown in Fig.~\ref{Fig:AnomalyTripletResult_2_normalize_test}. 
The accuracy for the test dataset improved significantly to 95.4\%.

From Fig.~\ref{Fig:AnomalyTripletResult_2_Distance} and Fig.~\ref{Fig:AnomalyTripletResult_2_T-SNE}, it can be seen that distance learning in the feature space is successfully performed, similar to Condition~1. 
By correctly learning distances in a two-digit dimensional feature space, the curse of dimensionality is avoided, resulting in improved estimation accuracy. 
In addition, Fig.~\ref{Fig:AnomalyTripletResult_2_Accuracy} and Fig.~\ref{Fig:AnomalyTripletResult_2_Loss} show that reducing the dimensionality of the feature space to 64 enables avoidance of overfitting without changing the number of epochs. 
This is considered to be because reducing the dimensionality of the feature space decreases the number of features learned by deep learning, thereby increasing the number of epochs required for appropriate learning.

\clearpage
Since reducing the dimensionality of the feature space was found to be effective for improving estimation accuracy, further experiments were conducted by reducing the dimensionality even more. 
In addition, when the dimensionality was reduced to 64, overfitting occurred earlier due to the reduced number of features. 
Therefore, the total number of training epochs was also increased.

\begin{figure}[h]
    \centering
    \includegraphics[width=0.7\linewidth]{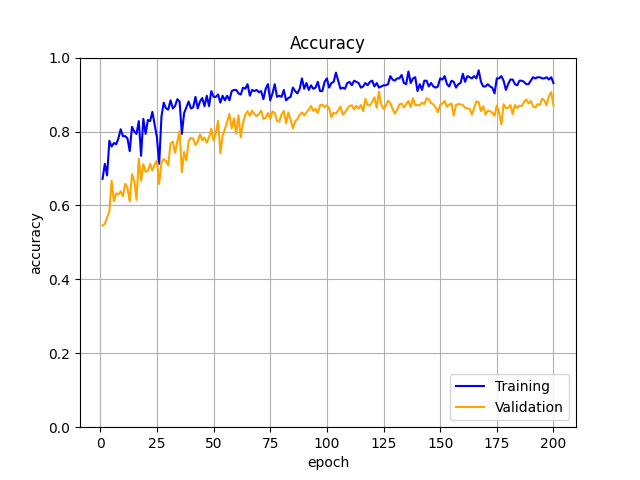}
    \caption{AnomalyTriplet Condition 3: Transition of accuracy with respect to the number of training epochs}
    \label{Fig:AnomalyTripletResult_3_Accuracy}
\end{figure}

\begin{figure}[h]
    \centering
    \includegraphics[width=0.7\linewidth]{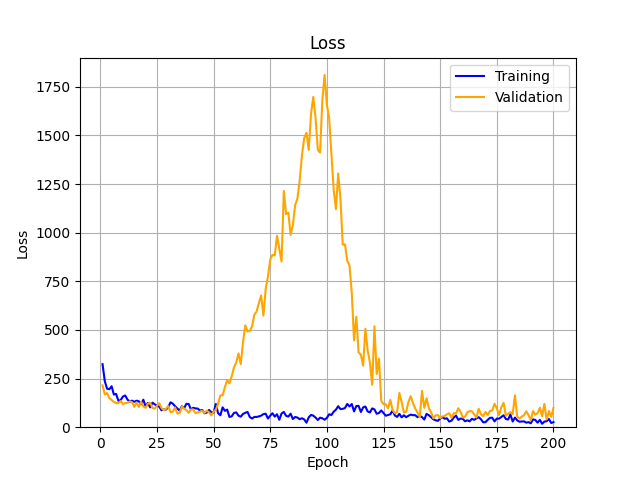}
    \caption{AnomalyTriplet Condition 3: Transition of loss with respect to the number of training epochs}
    \label{Fig:AnomalyTripletResult_3_Loss}
\end{figure}

\begin{figure}[h]
    \centering
    \includegraphics[width=0.7\linewidth]{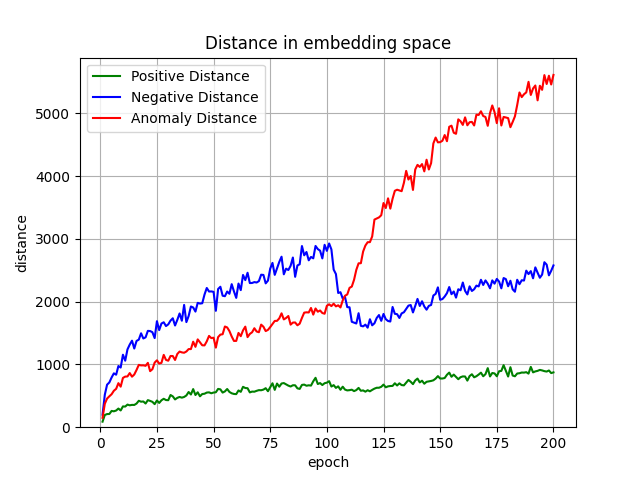}
    \caption{AnomalyTriplet Condition 3: Distances between samples in the feature space}
    \label{Fig:AnomalyTripletResult_3_Distance}
\end{figure}

\begin{figure}[h]
    \centering
    \includegraphics[width=0.7\linewidth]{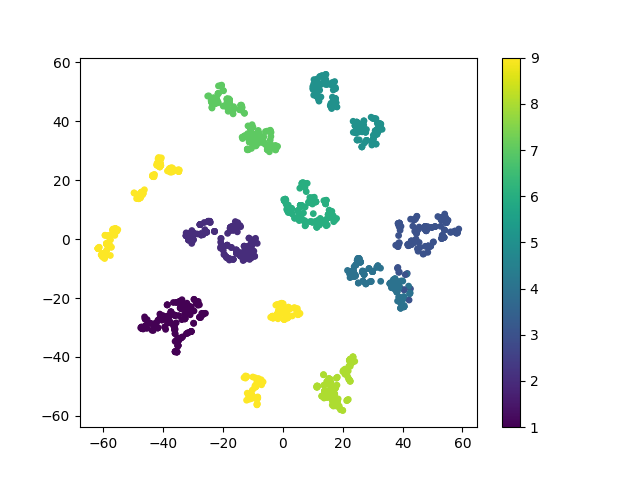}
    \caption{AnomalyTriplet Condition 3: Two-dimensional visualization of the feature space using t-SNE}
    \label{Fig:AnomalyTripletResult_3_T-SNE}
\end{figure}

\begin{figure}[h]
    \centering
    \includegraphics[width=0.7\linewidth]{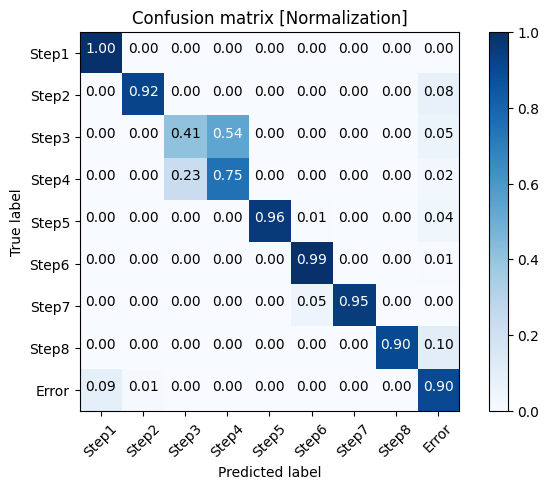}
    \caption{AnomalyTriplet Condition 3: Evaluation using a confusion matrix}
    \label{Fig:AnomalyTripletResult_3_normalize_test}
\end{figure}

The training conditions were set as follows: the total number of epochs was 200, anomaly learning started from epoch 100, $k=10$ for kNN, the dimensionality of the feature space was reduced to 32, and the learning rate was set to $lr=0.0001$. 
This condition is referred to as \textit{AnomalyTriplet Condition 3}. 
The results of training, including accuracy, loss, distances in the feature space, and two-dimensional visualization of the feature space using t-SNE, are shown in Fig.~\ref{Fig:AnomalyTripletResult_3_Accuracy}, Fig.~\ref{Fig:AnomalyTripletResult_3_Loss}, Fig.~\ref{Fig:AnomalyTripletResult_3_Distance}, and Fig.~\ref{Fig:AnomalyTripletResult_3_T-SNE}, respectively. 
The confusion matrix of the progress estimation results is shown in Fig.~\ref{Fig:AnomalyTripletResult_3_normalize_test}. 
The accuracy for the test dataset was 90.6\%.

Although Fig.~\ref{Fig:AnomalyTripletResult_3_Accuracy}, Fig.~\ref{Fig:AnomalyTripletResult_3_Distance}, Fig.~\ref{Fig:AnomalyTripletResult_3_T-SNE}, and Fig.~\ref{Fig:AnomalyTripletResult_3_normalize_test} suggest that learning proceeded successfully, Fig.~\ref{Fig:AnomalyTripletResult_3_Loss} shows that the validation loss rapidly worsens after approximately 50 epochs. 
This degradation is considered to be caused by the fact that anomaly learning starts from epoch 100. 
Until anomaly learning begins, the model is trained only with the conventional Triplet Loss. 
As a result, the Triplet Loss learning part becomes overfitted before anomaly learning starts, leading to a decrease in overall estimation accuracy.

\clearpage
To avoid overfitting while maintaining a feature space dimensionality of 32, experiments were conducted again with modified training conditions. 
To prevent overfitting, the total number of training epochs was reduced compared to AnomalyTriplet Condition~3.

\begin{figure}[h]
    \centering
    \includegraphics[width=0.7\linewidth]{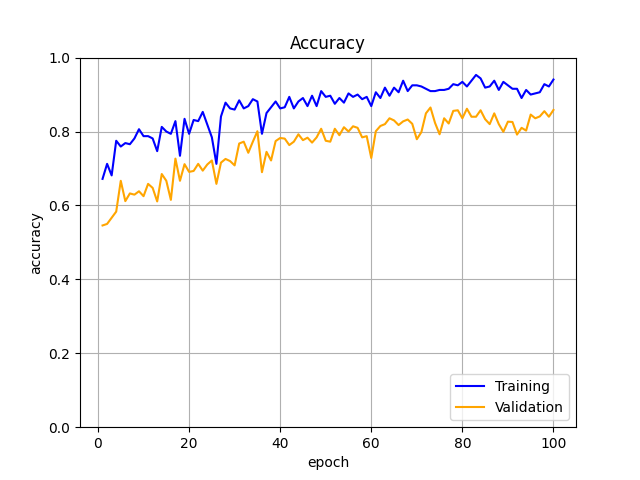}
    \caption{AnomalyTriplet Condition 4: Transition of accuracy with respect to the number of training epochs}
    \label{Fig:AnomalyTripletResult_4_Accuracy}
\end{figure}

\begin{figure}[h]
    \centering
    \includegraphics[width=0.7\linewidth]{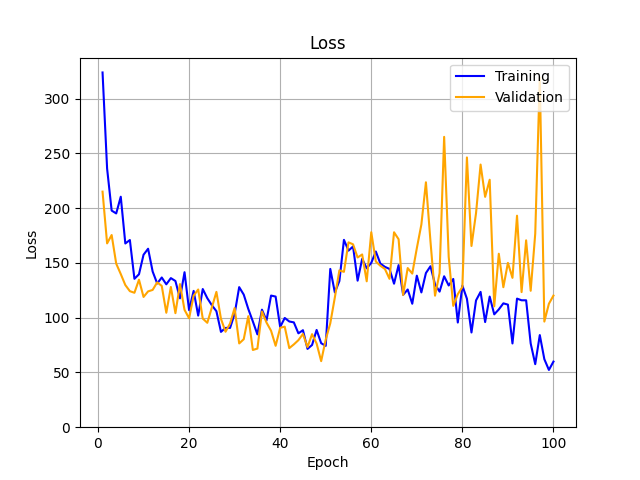}
    \caption{AnomalyTriplet Condition 4: Transition of loss with respect to the number of training epochs}
    \label{Fig:AnomalyTripletResult_4_Loss}
\end{figure}

\begin{figure}[h]
    \centering
    \includegraphics[width=0.7\linewidth]{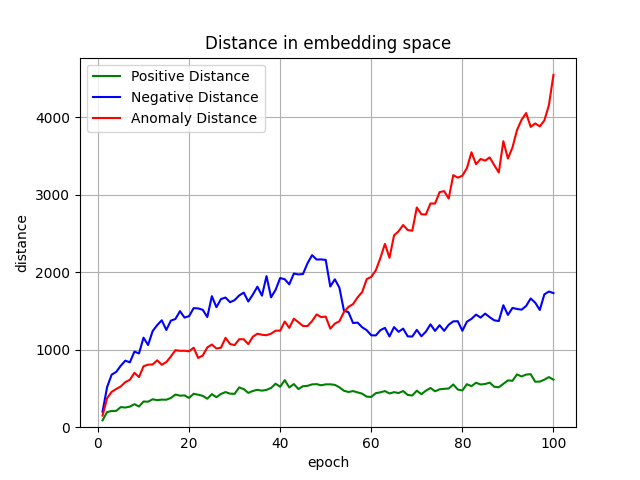}
    \caption{AnomalyTriplet Condition 4: Distances between samples in the feature space}
    \label{Fig:AnomalyTripletResult_4_Distance}
\end{figure}

\begin{figure}[h]
    \centering
    \includegraphics[width=0.7\linewidth]{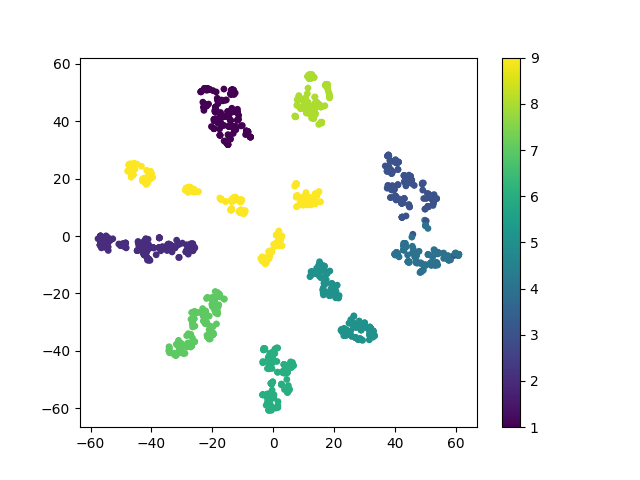}
    \caption{AnomalyTriplet Condition 4: Two-dimensional visualization of the feature space using t-SNE}
    \label{Fig:AnomalyTripletResult_4_T-SNE}
\end{figure}

\begin{figure}[h]
    \centering
    \includegraphics[width=0.7\linewidth]{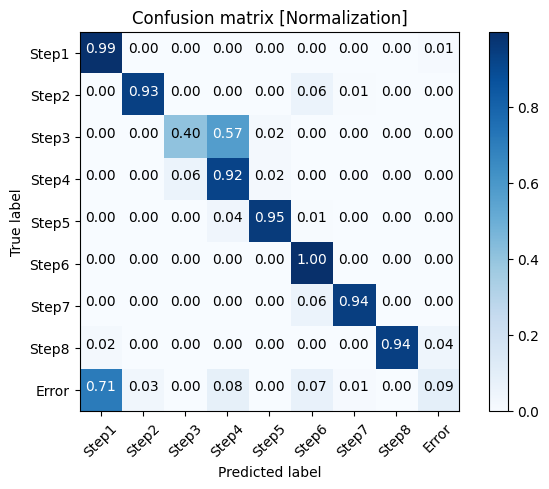}
    \caption{AnomalyTriplet Condition 4: Evaluation using a confusion matrix}
    \label{Fig:AnomalyTripletResult_4_normalize_test}
\end{figure}

The training conditions were set as follows: the total number of epochs was 100, anomaly learning started from epoch 50, $k=10$ for kNN, the dimensionality of the feature space was 32, and the learning rate was set to $lr=0.0001$. 
This condition is referred to as \textit{AnomalyTriplet Condition 4}. 
The results of training, including accuracy, loss, distances in the feature space, and two-dimensional visualization using t-SNE, are shown in Fig.~\ref{Fig:AnomalyTripletResult_4_Accuracy}, Fig.~\ref{Fig:AnomalyTripletResult_4_Loss}, Fig.~\ref{Fig:AnomalyTripletResult_4_Distance}, and Fig.~\ref{Fig:AnomalyTripletResult_4_T-SNE}, respectively. 
The confusion matrix of the progress estimation results is shown in Fig.~\ref{Fig:AnomalyTripletResult_4_normalize_test}. 
The accuracy for the test dataset was 76.3\%.

From Fig.~\ref{Fig:AnomalyTripletResult_4_Accuracy} and Fig.~\ref{Fig:AnomalyTripletResult_4_Loss}, it can be confirmed that overfitting does not occur before epoch 50, when anomaly learning begins. 
In addition, after epoch 50, the accuracy continues to improve until epoch 100, indicating that training is insufficient at 100 epochs and that further learning is required.

\clearpage
Finally, experiments were conducted by adjusting the number of epochs to avoid overfitting while ensuring sufficient learning with a feature space dimensionality of 32. 
The total number of epochs was increased to 200, while the anomaly learning start epoch was kept unchanged at epoch 50.

\begin{figure}[h]
    \centering
    \includegraphics[width=0.7\linewidth]{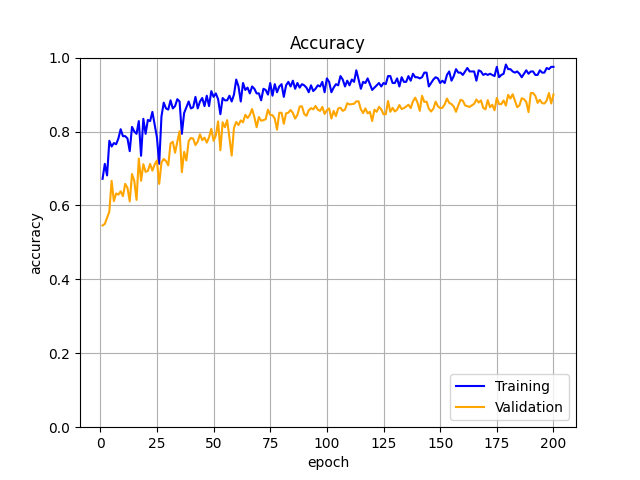}
    \caption{AnomalyTriplet Condition 5: Transition of accuracy with respect to the number of training epochs}
    \label{Fig:AnomalyTripletResult_5_Accuracy}
\end{figure}

\begin{figure}[h]
    \centering
    \includegraphics[width=0.7\linewidth]{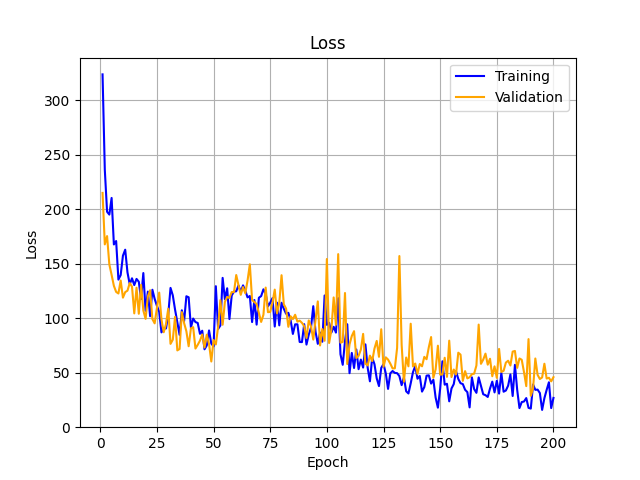}
    \caption{AnomalyTriplet Condition 5: Transition of loss with respect to the number of training epochs}
    \label{Fig:AnomalyTripletResult_5_Loss}
\end{figure}

\begin{figure}[h]
    \centering
    \includegraphics[width=0.7\linewidth]{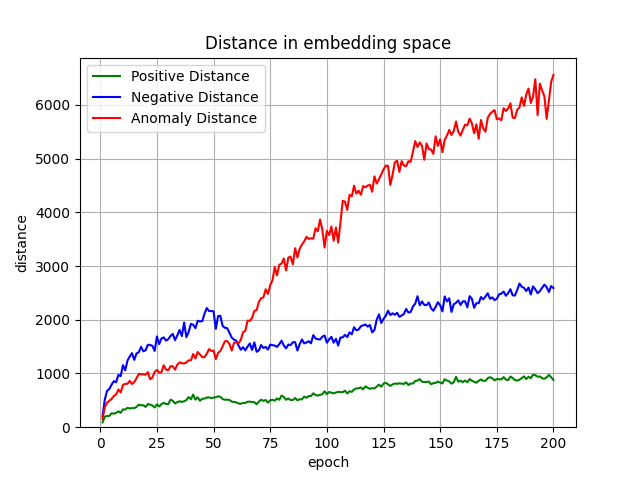}
    \caption{AnomalyTriplet Condition 5: Distances between samples in the feature space}
    \label{Fig:AnomalyTripletResult_5_Distance}
\end{figure}

\begin{figure}[h]
    \centering
    \includegraphics[width=0.7\linewidth]{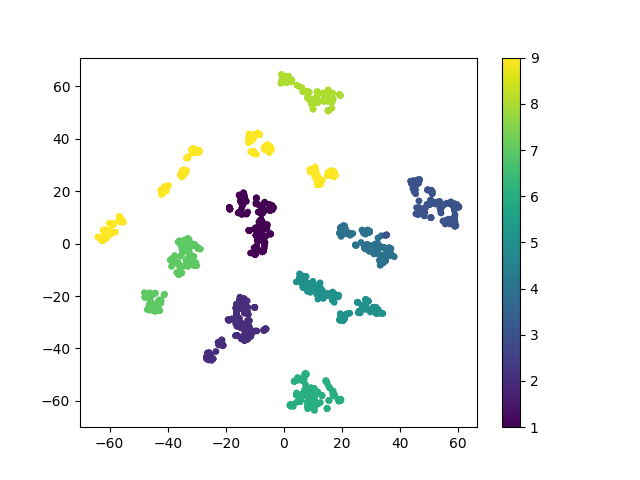}
    \caption{AnomalyTriplet Condition 5: Two-dimensional visualization of the feature space using t-SNE}
    \label{Fig:AnomalyTripletResult_5_T-SNE}
\end{figure}

\begin{figure}[h]
    \centering
    \includegraphics[width=0.7\linewidth]{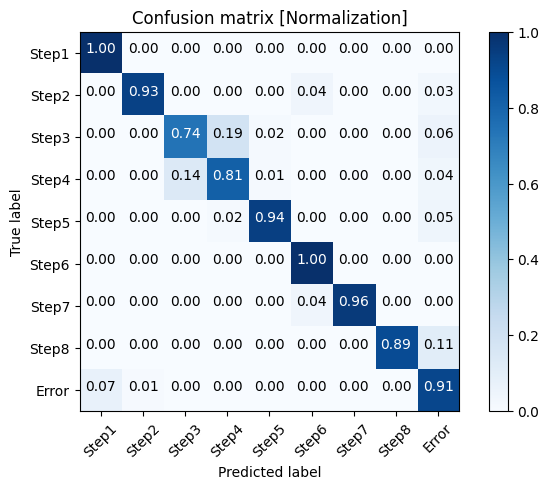}
    \caption{AnomalyTriplet Condition 5: Evaluation using a confusion matrix}
    \label{Fig:AnomalyTripletResult_5_normalize_test}
\end{figure}

The training conditions were set as follows: the total number of epochs was 200, anomaly learning started from epoch 50, $k=10$ for kNN, the dimensionality of the feature space was 32, and the learning rate was set to $lr=0.0001$. 
This condition is referred to as \textit{AnomalyTriplet Condition 5}. 
The results of training, including accuracy, loss, distances in the feature space, and two-dimensional visualization using t-SNE, are shown in Fig.~\ref{Fig:AnomalyTripletResult_5_Accuracy}, Fig.~\ref{Fig:AnomalyTripletResult_5_Loss}, Fig.~\ref{Fig:AnomalyTripletResult_5_Distance}, and Fig.~\ref{Fig:AnomalyTripletResult_5_T-SNE}, respectively. 
The confusion matrix of the progress estimation results when the error determination distance threshold was set to 1,025,000 is shown in Fig.~\ref{Fig:AnomalyTripletResult_5_normalize_test}. 
The accuracy for the test dataset reached 96.5\%.

From Fig.~\ref{Fig:AnomalyTripletResult_5_Accuracy} and Fig.~\ref{Fig:AnomalyTripletResult_5_Loss}, it can be confirmed that overfitting does not occur under this condition. 
Furthermore, Fig.~\ref{Fig:AnomalyTripletResult_5_Distance} and Fig.~\ref{Fig:AnomalyTripletResult_5_T-SNE} demonstrate that distance metric learning is correctly performed in the feature space.

Among the five experimental conditions evaluated in this section, AnomalyTriplet Condition~5 achieved the highest estimation accuracy. 
Therefore, in the subsequent sections, this condition is used as the baseline setting for training parameters, and comparative evaluations are conducted using this condition.

In Fig.~\ref{Fig:AnomalyTripletResult_5_normalize_test}, misclassification between Step~3 and Step~4, which was observed in earlier conditions, is reduced. 
Step~3 and Step~4 correspond to the states before and after installing the CPU onto the PC, respectively. 
Since the change in appearance features between these steps is small, misclassification is likely to occur. 
Under AnomalyTriplet Condition~5, the misclassification rate between adjacent steps was reduced to 4.8\%.

For reference, the total training time under this condition was 12.1 hours using an RTX2080 GPU.

To verify the effectiveness of the proposed Anomaly Quadruplet-Net, comparative experiments were conducted against the existing method, Anomaly Triplet-Net. 
The experiments were performed using a desktop PC assembly dataset. 
For comparison, the results of Anomaly Triplet-Net under AnomalyTriplet Condition~5 were used as the baseline.

A desktop PC was selected as the target assembly product. 
The dataset itself was identical to that used in the comparative experiments of Anomaly Triplet-Net. 
However, due to differences in the data loader design, the number of images used for training and testing was partially modified.

The training dataset remained unchanged and consisted of 40 images per step from Step~1 to Step~8, resulting in a total of 320 images. 
The validation dataset was also unchanged and consisted of 150 images per step, for a total of 1,264 images. 
The test dataset was modified in terms of the number of images; in addition to Step~1 through Step~8, error images were included, resulting in 129 images per step and a total of 1,161 images.

The experiments were conducted multiple times while varying the total number of epochs, the epoch at which anomaly learning was initiated, and the dimensionality of the feature space. 
The training parameters were determined empirically.

The training conditions were set as follows: the total number of epochs was 250, anomaly learning started from epoch 100, $k=10$ for the kNN algorithm, the dimensionality of the feature space was set to 64, and the learning rate was $lr=0.0001$. 
This setting is referred to as \textit{AnomalyQuadruplet Condition 1}.

\begin{figure}[h]
    \centering
    \includegraphics[width=0.7\linewidth]{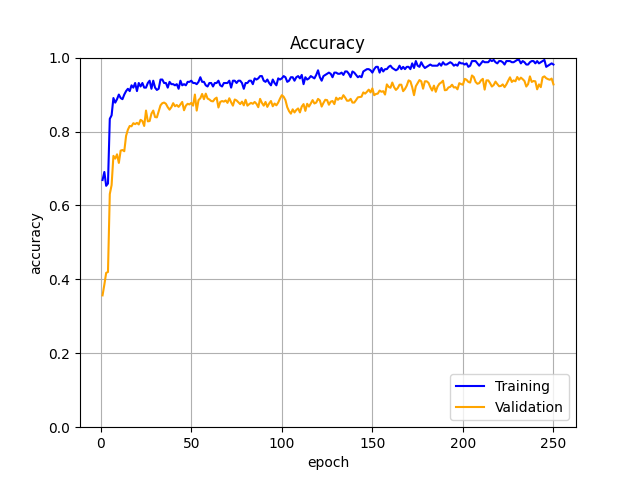}
    \caption{AnomalyQuadruplet Condition 1: Transition of accuracy with respect to the number of training epochs}
    \label{Fig:AnomalyQuadrupletResult_1_Accuracy}
\end{figure}

\begin{figure}[h]
    \centering
    \includegraphics[width=0.7\linewidth]{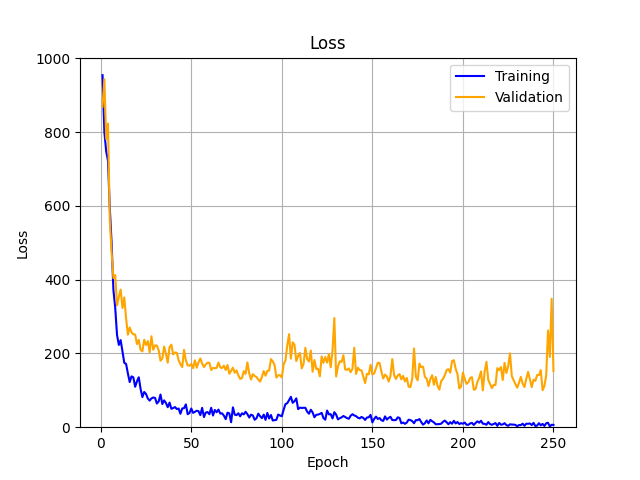}
    \caption{AnomalyQuadruplet Condition 1: Transition of loss with respect to the number of training epochs}
    \label{Fig:AnomalyQuadrupletResult_1_Loss}
\end{figure}

\begin{figure}[h]
    \centering
    \includegraphics[width=0.7\linewidth]{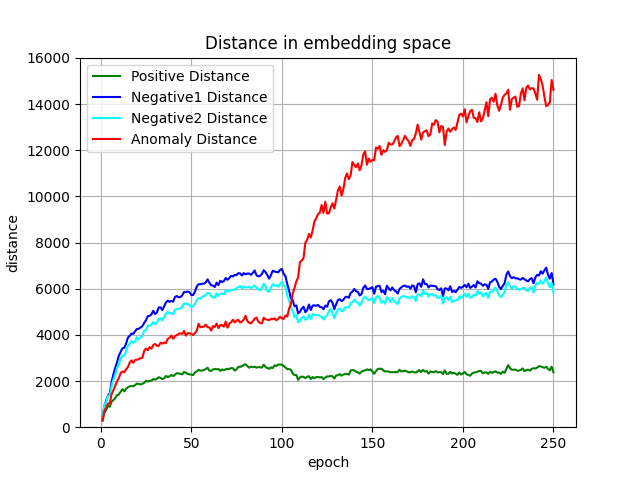}
    \caption{AnomalyQuadruplet Condition 1: Distances among samples in the feature space}
    \label{Fig:AnomalyQuadrupletResult_1_Distance}
\end{figure}

\begin{figure}[h]
    \centering
    \includegraphics[width=0.7\linewidth]{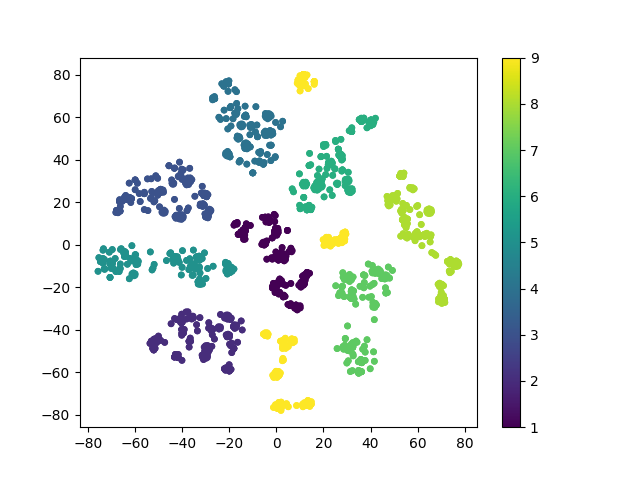}
    \caption{AnomalyQuadruplet Condition 1: Two-dimensional visualization of the feature space using t-SNE}
    \label{Fig:AnomalyQuadrupletResult_1_T-SNE}
\end{figure}

The training results, including the transition of accuracy, loss, distances in the feature space, and two-dimensional visualization of the feature space using t-SNE, are shown in Fig.~\ref{Fig:AnomalyQuadrupletResult_1_Accuracy}, Fig.~\ref{Fig:AnomalyQuadrupletResult_1_Loss}, Fig.~\ref{Fig:AnomalyQuadrupletResult_1_Distance}, and Fig.~\ref{Fig:AnomalyQuadrupletResult_1_T-SNE}, respectively.

\begin{figure}[h]
    \centering
    \includegraphics[width=0.7\linewidth]{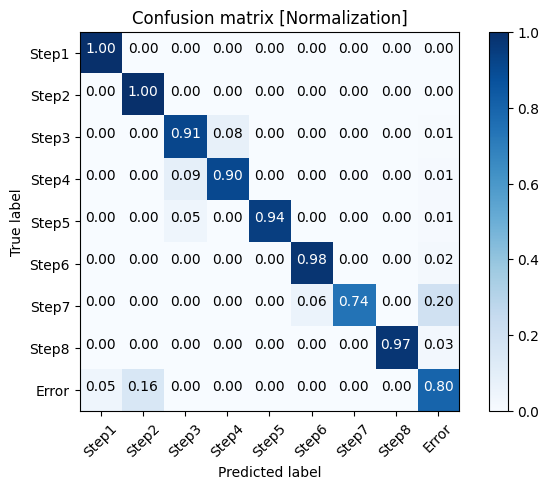}
    \caption{AnomalyQuadruplet Condition 1: Evaluation using a confusion matrix}
    \label{Fig:AnomalyQuadrupletResult_1_normalize_test}
\end{figure}

The confusion matrix of the progress estimation results, when the distance threshold for error determination was set to 9,000,000, is shown in Fig.~\ref{Fig:AnomalyQuadrupletResult_1_normalize_test}. 
The accuracy for the test dataset reached 97.8\%. 
The misclassification rate between adjacent steps was 2.9\%.

For reference, the total training time under this condition was 18.6 hours using an RTX4090 GPU.

From the experimental results, when applied to the desktop PC dataset, the proposed Anomaly Quadruplet-Net achieved an improvement of 1.3\% in accuracy compared to the conventional Anomaly Triplet-Net.
In addition, the misclassification rate between adjacent assembly steps was reduced by 1.9\%.

By comparing Fig.~\ref{Fig:AnomalyTripletResult_5_normalize_test} and Fig.~\ref{Fig:AnomalyQuadrupletResult_1_normalize_test}, it can be observed that misclassification was particularly reduced for the task of attaching the CPU to the PC between Step~3 and Step~4.
This task involves only subtle changes in appearance features, and the results confirm the effectiveness of the proposed Anomaly Quadruplet-Net, which was designed to reduce misclassification in such assembly processes where appearance changes are small.

\clearpage
\section{Conclusion}
\label{sec:conclusion:conc}

In this study, we developed an assembly progress estimation system that focuses on changes in the appearance features of assembly objects.
For progress estimation, we proposed a method based on deep metric learning.
In addition, to reduce misclassification in cases where changes in appearance features are small, we proposed Anomaly Quadruplet-Net, which utilizes Quadruplet Loss, and experimentally verified the effectiveness of the proposed methods.

Specifically, comparative experiments were conducted between the proposed Anomaly Quadruplet-Net and the prior method, Anomaly Triplet-Net.
The experiments were performed using a desktop PC dataset.
As a result, it was confirmed that Anomaly Quadruplet-Net achieved higher progress estimation performance than Anomaly Triplet-Net.

In the experiments using the desktop PC dataset, progress estimation was successfully performed with an accuracy of 97.8\%.
Furthermore, the misclassification rate between adjacent assembly steps was suppressed to 2.9\%.
Moreover, it was verified that the proposed method is particularly effective for adjacent assembly steps in which misclassification is likely to occur due to small changes in appearance features.

As future work, we plan to conduct experiments on a video-based progress estimation system using Anomaly Quadruplet-Net in environments where multiple objects are present.
In addition, in real environments, switching is expected to occur during object tracking.
Therefore, we aim to propose a system that utilizes ArUco markers and to implement a method that is practically applicable in industrial environments.

\bibliographystyle{unsrtnat}
\bibliography{references}  


\end{document}